\documentclass[10pt,twocolumn,letterpaper]{article}

\usepackage[pagenumbers]{cvpr} % To force page numbers, e.g. for an arXiv version

\usepackage{graphicx}
\usepackage{amsmath}
\usepackage{amssymb}
\usepackage{booktabs}

\usepackage{float}

\usepackage{times}
\usepackage{epsfig}
\usepackage{color}

\usepackage{stfloats} % added by author2

\usepackage{dsfont}
\usepackage{subcaption}
\usepackage{enumitem}
\usepackage{tabularx}
\usepackage{colortbl}
\usepackage{multirow}

\usepackage{marvosym}
\usepackage[export]{adjustbox}
\usepackage{pifont}

\definecolor{sh_gray}{rgb}{0.84,0.84,0.84}
\definecolor{sh_gray2}{rgb}{1,0.89,0.75}
\definecolor{color3}{rgb}{0.95,0.95,0.95}
\definecolor{color4}{rgb}{0.94,0.94,1}
\definecolor{color5}{rgb}{1,0.96,0.88}
\def\xnet{BiSTNet\xspace}

\newlength{\Oldarrayrulewidth}

\usepackage[pagebackref,breaklinks,colorlinks,citecolor=blue,linkcolor=red,bookmarks=false]{hyperref}

\usepackage[capitalize]{cleveref}
\crefname{section}{Sec.}{Secs.}
\Crefname{section}{Section}{Sections}
\Crefname{table}{Table}{Tables}
\crefname{table}{Tab.}{Tabs.}

\def\cvprPaperID{3826} % *** Enter the CVPR Paper ID here
\def\confName{CVPR}
\def\confYear{2023}

\begin{document}


% \title{\vspace{-0.5em}\xnet: A Progressively Dual Mask-Guided Network for Exemplar-Based Video Colorization\vspace{-0.5em}}

\title{\vspace{-0.5em}\xnet: Semantic Image Prior Guided Bidirectional Temporal Feature Fusion for Deep Exemplar-based Video Colorization\vspace{-0.5em}}

\author{
Yixin Yang$^{1}$ \quad Zhongzheng Peng$^{1}$ \quad Xiaoyu Du$^{1}$ \quad Zhulin Tao$^{2}$ \quad  Jinhui Tang$^{1}$  \quad Jinshan Pan$^{1}$ \\
$^1$Nanjing University of Science and Technology \\
$^{2}$ Communication University of China\\
\vspace{-0.5em}
}

% v0
\maketitle

\begin{figure}[t]\footnotesize
	\vspace{-3.3in}
	\begin{minipage}{\textwidth}
	\centering
	 \begin{tabular}{cccccccc}
	 \multicolumn{3}{c}{\multirow{5}*[58pt]{\includegraphics[width=0.368\linewidth, height=0.285\linewidth]{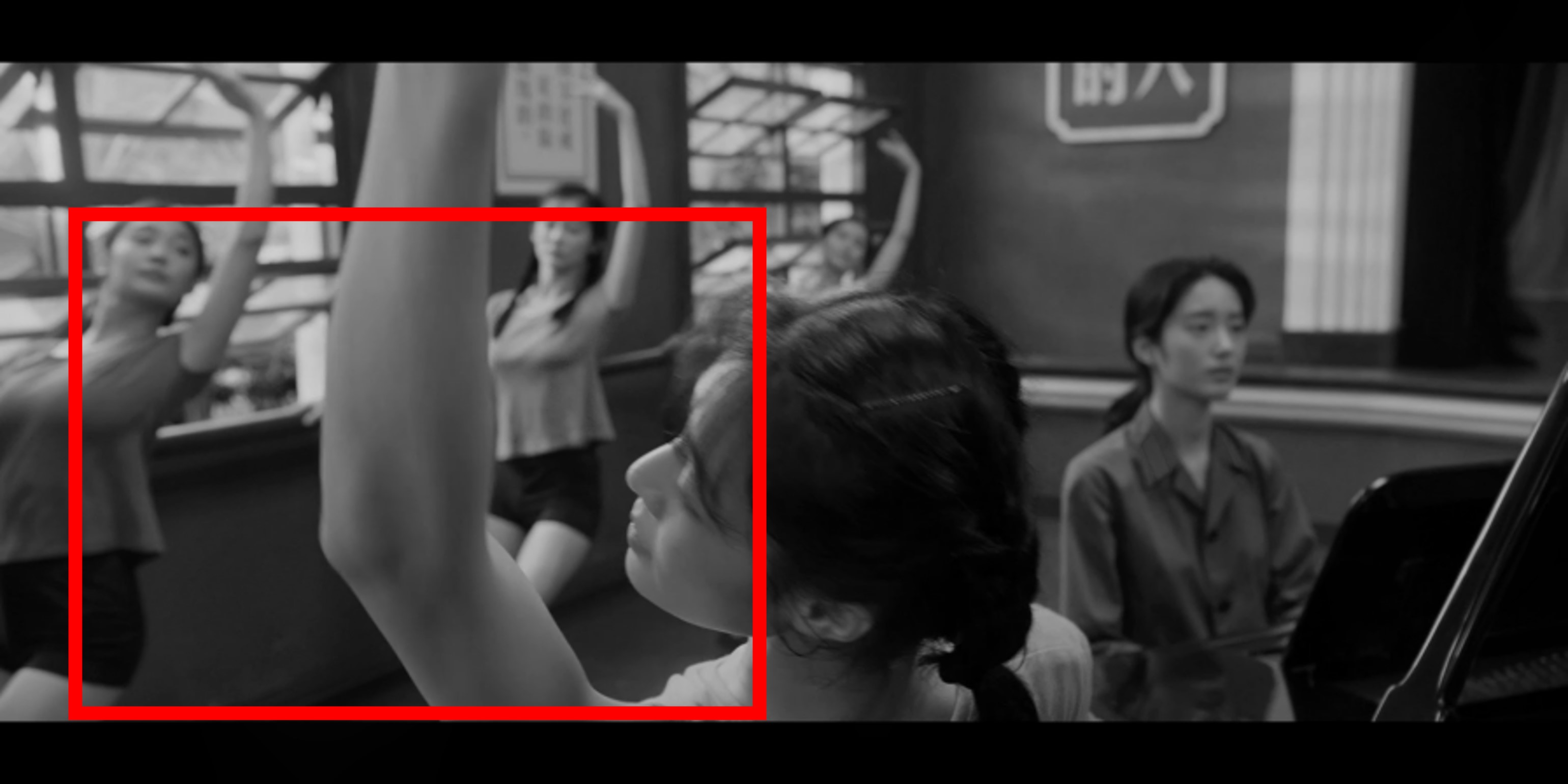}}}&\hspace{-3.5mm}
	 \includegraphics[width=0.14\linewidth, height = 0.127\linewidth]{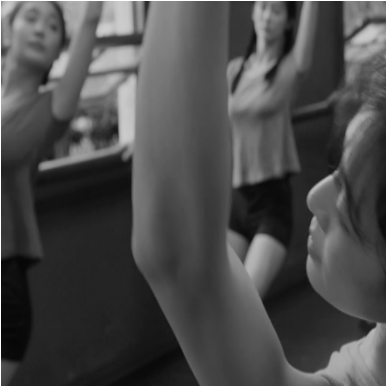} &\hspace{-3.5mm}
     \includegraphics[width=0.14\linewidth, height = 0.127\linewidth]{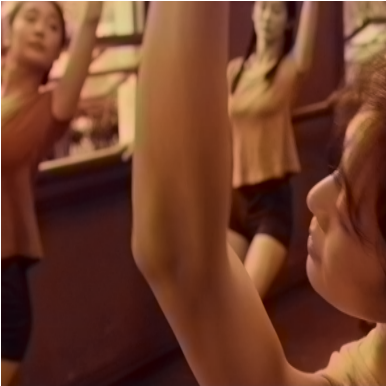} &\hspace{-3.5mm}
     \includegraphics[width=0.14\linewidth, height = 0.127\linewidth]{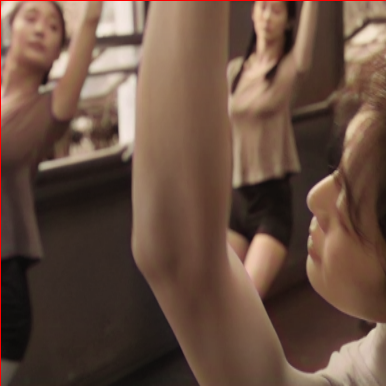} &\hspace{-3.5mm}
     \includegraphics[width=0.14\linewidth, height = 0.127\linewidth]{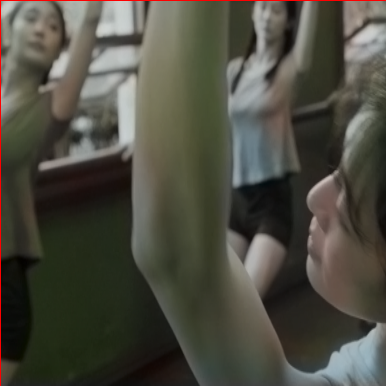} \\
	 \multicolumn{3}{c}{~} &\hspace{-3.5mm}  (a) Input &\hspace{-3.5mm}  (b) CIC~\cite{cic} &\hspace{-3.5mm}  (c) FAVC~\cite{favc}  &\hspace{-3.5mm}  (d) DeepRemaster~\cite{IizukaSIGGRAPHASIA2019}\\
     \multicolumn{3}{c}{~} & \hspace{-3.5mm}
     \includegraphics[width=0.14\linewidth, height = 0.127\linewidth]{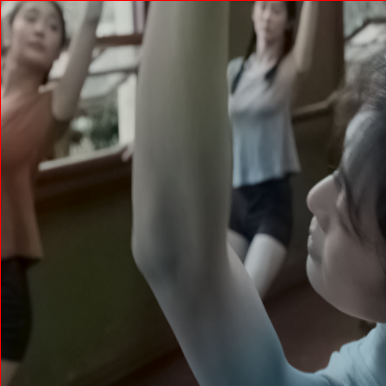} & \hspace{-3.5mm}
     \includegraphics[width=0.14\linewidth, height = 0.127\linewidth]{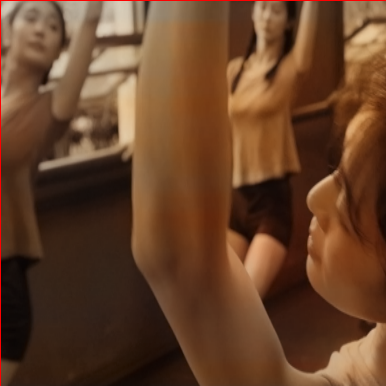} & \hspace{-3.5mm}
     \includegraphics[width=0.14\linewidth, height = 0.127\linewidth]{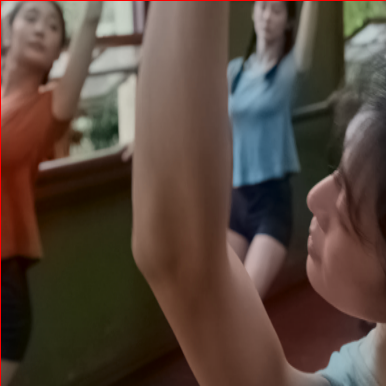} & \hspace{-3.5mm}
     \includegraphics[width=0.14\linewidth, height = 0.127\linewidth]{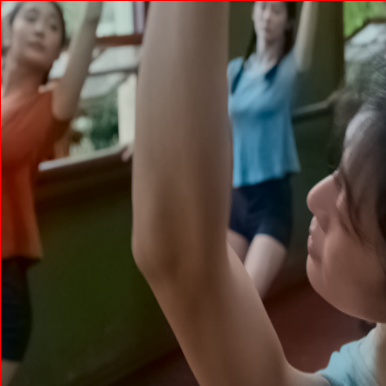} \\
  \multicolumn{3}{c}{\hspace{-3.5mm} Input frame from a real-world film \textit{Youth}} &\hspace{-3.5mm}  (e) DeepExemplar~\cite{zhang2019deep} &  \hspace{-3.5mm} (f) TCVC~\cite{liu2021temporally}  &\hspace{-3.5mm}  (g) Ours  & \hspace{-3.5mm}(h) Ground Truth \\
	\end{tabular}
	%\vspace{-0.02mm}
	\caption{
		Colorization results in a real-world challenging film \textit{Youth}. Our method explores well-colorized pixels from reference exemplars by a semantic correspondence and adaptively propagates the estimated colors by a bidirectional temporal feature fusion under the guidance of the semantic image prior to achieve better video colorization. As our analysis shows, reducing the influence of the inaccurately matched colors with the proposed bidirectional temporal feature fusion module and utilizing semantic information to guide the colorization process can obtain better-colorized results than both the exemplar-based methods~\cite{zhang2019deep, IizukaSIGGRAPHASIA2019} and the automatic colorization methods~\cite{cic,liu2021temporally,favc} in terms of accuracy and temporal consistency.
	}
	\label{fig:visual_results}
	\end{minipage}
	\vspace{-2mm}
\end{figure}

% % 1104 fanghua teaser
% \begin{figure}[t]\footnotesize
% \vspace{-1.2in}
% \begin{minipage}{\textwidth}
% \centering
% \begin{tabular}{ccccc}
% \includegraphics[width=0.19\linewidth]{images/fanghua292_abstract/00008_LR_Orange.png}&\hspace{-4.5mm}
% \includegraphics[width=0.19\linewidth]{images/fanghua292_abstract/00008_output_1020_CIC_original_cvpr23_eccv16_Orange.png} &\hspace{-4.5mm}
% \includegraphics[width=0.19\linewidth]{images/fanghua292_abstract/00008_output_0930_tcsvt_0712_v0_deepexemplar_baseline_Orange.png} &\hspace{-4.5mm}
% \includegraphics[width=0.19\linewidth]{images/fanghua292_abstract/00008_output_1103_DeepRemaster_Orange.png} &\hspace{-4.5mm}
% \includegraphics[width=0.19\linewidth]{images/fanghua292_abstract/00008_output_0930_tcsvt_0712_v3_videoInterp_DavisBatch_Orange.png} \\
% (a) Input frame & \hspace{-0.5cm} (b) CIC~\cite{cic} & \hspace{-0.5cm} (c) DeepExemplar~\cite{zhang2019deep} & \hspace{-0.5cm} (d) DeepRemaster~\cite{IizukaSIGGRAPHASIA2019} & \hspace{-0.5cm} (e) Ours\\
% \end{tabular}
% \vspace{-2mm}
% \caption{Colorized result on a real challenging video.
% %All the compared results are obtained from the reported results.
% %
% }
% \label{fig:teaser}
% %
% \end{minipage}
% %\vspace{-2mm}
% \end{figure}
% %\vspace{2mm}

%%%%%%%%% ABSTRACT
%\vspace{-5mm}
\begin{abstract}
%\vspace{-0.5em}

% 1027 v1

%We present an effective end-to-end network for exemplar-based video colorization. Instead of performing video colorization in one-way sequence, our method aggregates semantic correspondence calculated from two dissimilar reference images and adopt a temporal-related fusion mechanism to fuse the bidirectional propagated colors that are weighted by time distances with respect to the reference images, leveraging significant information about occlusion, to guide the colorization of every frame, which is critical to achieve temporal consistency and alleviate annoying artifacts. To further enhance visual performance, we introduce a mixed expert block to extract image prior like object boundaries or intensity changes to constrain colorization period reasonable and smooth, a scale recurrent block to gradually restore the colorful image in different resolutions within a image pyramid through “coarse-to-fine” scheme to improve frame-level performance. Extensive experiments have demonstrated that the proposed algorithm performs favorably against state-of-the-art methods on the benchmark datasets as well as real-world videos.

How to effectively explore the colors of reference exemplars and propagate them to colorize each frame is vital for exemplar-based video colorization.
In this paper, we present an effective \xnet to explore colors of reference exemplars and utilize them to help video colorization by using a bidirectional temporal feature fusion with the guidance of
semantic image prior.
We first establish the semantic correspondence between each frame and the reference exemplars in deep feature space to explore color information from reference exemplars.
Then, to better propagate the colors of reference exemplars into each frame and avoid the inaccurate matches colors from exemplars
we develop a simple yet effective bidirectional temporal feature fusion module to better colorize each frame.
We note that there usually exist color-bleeding artifacts around the boundaries of the important objects in videos. To overcome this problem, we further develop a mixed expert block to extract semantic information for modeling the object boundaries of frames so that the semantic image prior can better guide the colorization process for better performance.
In addition, we develop a multi-scale recurrent block to progressively colorize frames in a coarse-to-fine manner.
Extensive experimental results demonstrate that the proposed \xnet performs favorably against state-of-the-art methods on the benchmark datasets. Our code will be made available at \url{https://yyang181.github.io/BiSTNet/}.

\end{abstract}

%%% 1025
% 逻辑点：1. 为了解决单帧存在的问题，一定有一个衰减的过程， 所以提出多帧。
% 逻辑点：2. 溢色 失色 用专家模型，提供边缘信息，提供语义分割信息来解决
% 逻辑点：3. 整体全局与局部信息均衡统一，提出了coarse-to-fine的模型

% 有参考帧的上色网络
% 定理1：有相似度才能进行有价值的参考帧上色
% 定理2：在较长的视频序列中，单帧参考帧与不同时刻的视频帧相似度有高有低
% 定理3：如果只用单帧参考帧，在进行上色的过程中，随着相似度的变化，上色效果一定会有一定程度的衰减
% 基于以上观察，我们提出了双参考帧的方法

% !一定要记住，我们做的是  ！！！基于样例！！！ 的视频上色
% 对比的结果一定要加上DeepRemaster

%%% 1026
% 视频上色的分类
% 1. 评价指标要 主观+客观
% 两个模块：temporal consistency   用CDC客观评价+ 溢色失色问题 用PSNR即可

% 2. introduction要分段， 阅读性，逻辑性，
% 第二段：
% 视频上色都有什么方法，全自动+基于样例两类，存在什么样的问题，而不是说该方法不好
% TCVC存在什么问题， 介绍
% 最后都要总结一下，所以需要有什么样的方法来解决这些问题

% 3. 图一定要挑最好的
% 主观的不要写，尽量客观
% 解决问题要加原因      原因一定要有解释原因
%  谦虚  有竞争力的模型即可， 不要用极端的词汇

% 增加两篇文章的方法：
% 1. Blind Video Temporal Consistency via Deep Video Prior
% 2. Stylization-Based Architecture for Fast Deep Exemplar Colorization
% 3. Deep Remaster

% 格式上的问题， 不要在最后留一两个单词，容易给人不充实的感觉

%%%%%%%%% BODY TEXT
% \vspace{2mm}
\section{Introduction}%\vspace{-0.5em}
\label{sec:intro}

Video colorization is a problem of generating fully colorized videos from their grayscale (monochrome) version. There are amounts of legacy black-and-white movies captured in the past decades due to the low quality of the film technology at the time, this problem has attracted lots of attention recently.
Video colorization is an ill-posed problem as only monochrome videos are given.
Compared to the single image colorization problem, the video colorization problem is more challenging as it not only needs to restore high-quality frames but also requires colorized videos with better temporal consistency.

% 先讲全自动的方法
The simplest way to solve video colorization is to process each frame independently using an image-based colorization model. Recently, amounts of single-image colorization methods are proposed and have achieved remarkable progress \cite{Levin2004,Cheng2015,cic,Larsson2016,let,tog17,inscolor}. However, these methods often lead to temporal flickering artifacts and discontinuities.
Thus, the demand for high-quality colorized videos, that look natural and are free of flickering artifacts, brings a huge challenge for existing video colorization methods.
% to bring high quality videos which means to be natural looking and free of flickering artifacts specific designed video colorization methods are desired

To constrain the temporal consistency in video colorization tasks, several automatic colorization methods are proposed, including utilizing self-regularization and temporal constraint \cite{favc}, conditional generative adversarial networks (GANs) \cite{8781608} and bidirectional deep feature propagation \cite{liu2021temporally}. These methods mainly focus on constraining temporal consistency while the colorization performance of each frame is far from satisfactory.
Moreover, as video colorization is ill-posed, only depending on the training data does not provide enough information for colorization.
Therefore, another category of approaches, exemplar-based colorization methods attract more attention recently.
%
% Therefore, another category of work, exemplar-based colorization which colorizing the grayscale images by transferring the color from the reference image in a similar content attracts amounts of attention recently.
%

Exemplar-based video colorization methods \cite{Xu_2020_CVPR, zhang2019deep, wan2022oldfilm, IizukaSIGGRAPHASIA2019} usually transfer the colors from a reference exemplar image to the grayscale one. Compared with automatic methods, exemplar-based methods use the color information of the reference exemplar to colorize every frame within the entire colorization period and are flexible enough to generate any kind of style according to the provided  reference exemplar images.
%
%\cite{zhang2019deep } proposes the first end-to-end network for exemplar-based video colorization consisting of two major modules: a correspondence subnet to capture similarities between the reference and the input frame, and a colorization subnet to refine coarsely warped color. By taking account information from multiple frames of the input video at once, \cite{IizukaSIGGRAPHASIA2019} develops an attention based network with temporal source-reference attention block to enhance visual performance. To achieve better spatial-temporal consistency, \cite{wan2022oldfilm} makes full use of hidden knowledge learned from adjacent frames through bidirectional propagation to obtain visual performance while ensuring temporal coherency.
%
These methods usually work well on short video clips or simple scenes, but the visual performance of colorized videos is far from satisfactory when handling long or complicated real-world scenes. Because the semantic objects within a single reference image usually cannot cover all objects shown up in the video clips, causing wrong colors to be transferred from the reference exemplar image to the input frame.
%
% The visual performance decreasing gradually when far away from the reference.
%
Thus, to design an effective video colorization network, it is essential to enhance the temporal consistency and maintain colorization performance for individual frames at the same time.

In this paper, we present an effective exemplar-based deep video colorization method.
To better explore the colors of reference exemplars and propagate them to help colorize each frame, we first establish the semantic correspondence between each frame and the reference exemplars in a deep feature space. However, we note that the inaccurately matched colors would affect colorization. In contrast to existing exemplar-based video colorization methods that directly use the match colors to colorize each frame, we develop a simple yet effective bidirectional temporal feature fusion to avoid the influence of the possible inaccurate estimated colors and better propagate the colors extracted from reference exemplars.
In addition, we note that most existing ones usually suffer from the color-bleeding artifact, a problematic color spreading near the boundaries of some important objects in videos.
To overcome this problem, we explore the semantic information of images to localize the boundaries of objects so that they can better guide the colorization for the regions of the object boundaries.
We formulate the proposed method into a multi-scale recurrent framework that can progressively colorize frames in a coarse-to-fine manner.
Both quantitative and qualitative evaluations show the effectiveness of the proposed method performs favorably against state-of-the-art ones (see Figure~\ref{fig:visual_results}).

% In this work, we propose \xnet, an effective multi-scale mixed colorization network, which aims at decreasing spatial-temporal inconsistency and color bleeding or color spilling problem for video colorization. It inherits the advantages in reference based methods like DeepExemplar, but also aims to improve the performance with the power of FusionModel and ExpertModels which would be discussed in the future section.

The main contributions of this work are summarized below:
\begin{itemize}[leftmargin=*]\setlength{\itemsep}{-0.2em}
    %  \item We present the end-to-end network for exemplar-based video colorization. We are the first
    % to use two reference frames, exactly the start and the end frame
    % of video clips.
     \item We present an effective method that explores both spatial information and temporal clues to keep the useful color information of provided exemplars for video colorization.
    \item We propose a bidirectional temporal fusion block (BTFB)  that is capable of combining bidirectional semantic correspondence based on temporal clues.
    This module is effective to avoid abrupt flicking between adjacent frames and reduce the influence of inaccurate colors from the reference exemplars for better colorization.
    % achieve temporal consistency while remaining
    % faithful to the reference style, colorization results from a single-
    % exemplar colorization network based on each reference frame are
    % fused properly by their temporal position.
    \item To ease color bleeding or color spilling effects, we develop a mixed expert block (MEB) to explore the semantic information and edge information as prior. By utilizing such image prior, MEB effectively guides the colorization of the regions around the object boundaries.
    \item We formulate the proposed method into a multi-scale recurrent convolutional block (MSRB) to progressively restore colorful video frames in a coarse-to-fine manner.
\end{itemize}
We extensively evaluate the proposed method on both synthetic datasets and real-world videos and demonstrate that it generates better-colorized videos in terms of accuracy and temporal consistency.

\section{Related Work}
\label{sec:background}

\noindent \textbf{Interactive colorization.} In the early stage of colorization, local user hints are the most popular~\cite{Levin2004,Qu2006,Luan2007,Chen2012,Yatziv2006}. These methods require input from the user either in the form of points, strikes, or scribbles and rely on the assumption that nearby pixels should have similar colors. Levin et al.~\cite{Levin2004} propose an interactive colorization technique that propagates colors from scribbles to neighboring similar pixels. Sangkloy et al.~\cite{Sangkloy_2017_CVPR} use an end-to-end feed-forward deep generative adversarial architecture to colorize images. To guide structural information and color patterns, user input in the form of sketches and color strokes is employed. Zhang et al.~\cite{tog17} develop user interaction based on two variants, local hint and global hint networks, both of which utilize a common main branch for image colorization. The non-trivial human effort and aesthetic skills to generate colorful images required by user-guided methods make them unsuitable for video colorizing tasks.

\noindent \textbf{Exemplar-based colorization.}
Exemplar-based colorization controls the color styles of the output frames by using reference frames. Deep exemplar-based colorization~\cite{he2018deep, zhang2019deep} aims to provide diverse colors to the same image. The system is composed of two subnetworks: both the similarity sub-network and the colorization sub-network. Inspired by stylization characteristics in feature extracting and blending, Xu et al.~\cite{Xu_2020_CVPR} propose a stylization-based architecture for fast deep exemplar colorization to reduce high time and resource consumption. To colorize the images with multiple objects, Su et al.~\cite{inscolor} propose an instance-aware image colorization method by utilizing an off-the-shelf pre-trained model to detect object instances and produce cropped object images. In those works, the correspondence and the color propagation are optimized independently, therefore visual artifacts tend to arise due to correspondence errors.
% Another category of work colorizes the grayscale images by transferring the color from the reference image in a similar content. The pioneering work transfers the chromatic information to the corresponding regions by matching the luminance and texture. In order to achieve a more accurate local transfer, various correspondence techniques have been proposed by matching low-level hand-crafted features. Still, these correspondence methods are not robust to complex appearance variations of the same object because low-level features do not capture semantic information. More recent works rely on the Deep Analogy method to establish the semantic correspondence and then refine the colorization by solving Markov random field model or a neural network. In those works, the correspondence and the color propagation are optimized independently, therefore visual artifacts tend to arise due to correspondence error. On the contrary, we unify the two stages within one network, which is trained end-to-end and produces more coherent colorization results.

\noindent \textbf{Video colorization.}
Video colorization~\cite{favc, paul2016spatiotemporal, sheng2013video, Xu_2020_CVPR, zhang2019deep} needs to consider both colorization performance and temporal consistency. Zhang et al.~\cite{zhang2019deep} unify the semantic correspondence and colorization into a single network and train it end-to-end to produce temporal consistent video colorization with realistic effects. Iizuka et al. propose DeepRemaster~\cite{IizukaSIGGRAPHASIA2019}, a fully 3D convolutional method, utilizing source-reference attention to colorize long videos without the need for segmentation while maintaining temporal consistency. Recently, FAVC~\cite{favc} is proposed for automatic video colorization. This method regularizes its model with a KNN graph built on the ground-truth color video and simultaneously posed a temporal loss term for constraining temporal consistency. FAVC focuses mainly on temporal consistency. Lai et al.~\cite{Lai2018} propose an approach to enforce stronger temporal consistency of a video generated frame by frame by an image processing algorithm such as colorization.

 \begin{figure*}[!t]
    \centering
    \includegraphics[width=0.95\textwidth,valign=t]{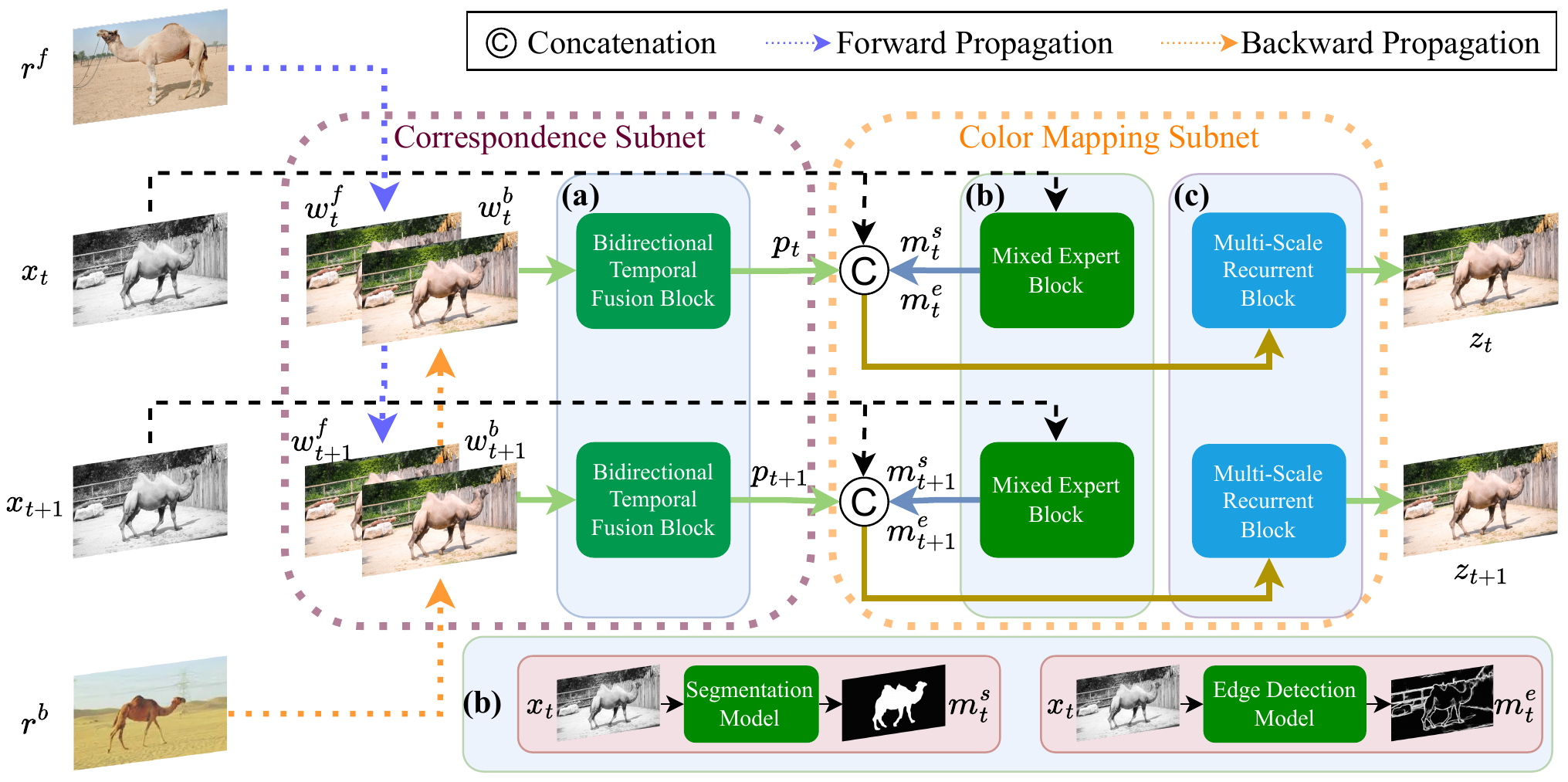}
    \vspace{-2mm}
    \caption{The architecture of \xnet for exemplar-based video colorization. The core components of our method include: \textbf{(a)} bidirectional temporal fusion block (BTFB), \textbf{(b)} mixed expert block (MEB) and \textbf{(c)} multi-scale recurrent block (MSRB).}
    \label{fig:framework}
    %\vspace{-5mm}
\end{figure*}

%\vspace{-0.5em}
\section{Proposed Method}\label{sec:method}
% Our main goal is to develop an end-to-end CNN model for exemplar-based video colorization that aims at effectively handle real-world relatively long videos and the spatial-temporal consistency constrain is highly demanded.
% Our main goal is to develop an effective end-to-end CNN model that can handle real-world relatively long videos for exemplar-based video colorization tasks.
Our main goal is to develop an effective model that explores both spatial and temporal information to generate colorful videos with better temporal consistency.
To this end, we introduce key designs such as, a bidirectional temporal fusion block employed to avoid the influence of inaccurate color information from reference exemplars and alleviate flicking artifacts, a mixed expert block to extract semantic image prior to better guide the colorization of the regions of object boundaries and a multi-scale recurrent block that is capable of performing color mapping in coarse-to-fine mechanism than a single-scale network \cite{zhang2019deep, Xu_2020_CVPR}, which benefits the single-frame colorization performance.

In this section, we first present the overall pipeline of our video colorization network architecture in detail. Then we describe the core components of the proposed blocks: (a) bidirectional temporal fusion block (b) mixed expert block and (c) multi-scale recurrent block. Finally, we introduce the loss functions.

% We aim to develop an efficient large-kernel CNN model for the SISR task. To meet the efficiency goal, we introduce key designs to the feature mixing block employed to encode information efficiently. This section first presents the overall pipeline of our proposed ShuffleMixer network in detail. Then, we formulate the feature mixing block, which acts as a basic module for building the ShuffleMixer network. Finally, we provide detail on the training loss function.

\noindent \textbf{Overall pipeline.}
%The target of video colorization is to generate fully colorized videos from its grayscale (monochrome) version. We perform this task in CIE $Lab$ color space, where $l$ represent the luminance in $Lab$ color space and $ab$ represent the chrominance. Thus, the objective of video colorization is to predict two associated chrominance channels of a grayscale image, \textit{i.e.}, $a$ channel and $b$ channel.
%
Given a grayscale video $X = \{x_0, x_1,...,x_{N-1} \}$, where $x_{t}$~$\in$~$\mathbb{R}^{H\times W \times 1}$ denotes the frame at time $t$, $H \times W$ denotes the spatial resolution, our main goal is to estimate the colorized video frames $Z = \{z_0, z_1,...,z_{N-1} \}$, where $z_{t}$~$\in$~$\mathbb{R}^{H\times W \times 3}$.
% Different from past works either using arbitrary number of reference color images concatenated together to pass through source-reference attention layer\cite{IizukaSIGGRAPHASIA2019} or using single reference color image acquired by inquiring the most similar image from the corresponding class in public datasets \cite{zhang2019deep}, here
%
% We use two reference color images from each video clip. The first reference color image at time $t=0$ is denoted as $r^{lab}_{0}$~$\in$~$\mathbb{R}^{H\times W \times 3}$ and the reference image at time $t=N-1$ is denoted as $r^{lab}_{N-1}$
%
%In preprocessing period, we colorize the first input frame $x_0$ and the last frame $x_{N-1}$ to obtain the reference color images, denoting them as $r^{f}$~$\in$~$\mathbb{R}^{H\times W \times 3}$ and $r^{b}$~$\in$~$\mathbb{R}^{H\times W \times 3}$, respectively.
%
% We use the ground truth color frames at time $0$ and time $N-1$ as the reference images $r^{f}$~$\in$~$\mathbb{R}^{H\times W \times 3}$ and $r^{b}$~$\in$~$\mathbb{R}^{H\times W \times 3}$ for one video clip.
%
We use two reference images $r^{f}$~$\in$~$\mathbb{R}^{H\times W \times 3}$ and $r^{b}$~$\in$~$\mathbb{R}^{H\times W \times 3}$ for each video clip. $r^{f}$ and $r^b$ contain similar contents to the input frames at time $0$ and time $N$, respectively.
%
%We use $f$ to represent that this reference image will be used in forward propagation and $b$ to represent backward propagation.
%
%We use the ground truth color images at time $0$ and time $N-1$ as reference images for one video clip in our experiments.
%Then, our \xnet uses a two-stage network which mainly consists of a correspondence subnet and a color mapping subnet to transfer the colors of these two reference frames.
%

Specifically, we first apply the correspondence subnet that contains a forward stage and a backward stage to calculate bidirectional semantic correspondence given input grayscale image $x_t$.
The forward semantic correspondence is to obtain the relationship between  $x_t$ and $r^{f}$ and the backward semantic correspondence is to obtain the relationship between $x_t$ and $r^{b}$.
We establish these semantic correspondence using the deep features $F_x^t$, $F_r^f$ and $F_r^b \in \mathbb{R}^{H\times W \times C}$ that are extracted from $x_t$, $r^{f}$ and $r^{b}$ by the pre-trained VGG19 model~\cite{vgg2014}. Then we reshape these features into feature vectors $\mathbf{F}_x^t$, $\mathbf{F}_r^f$ and $\mathbf{F}_r^b \in \mathbb{R}^{HW \times C}$.
%$F_x$~$\in$~$\mathbb{R}^{H\times W \times C}$ is the feature for input frame $x_t$ with $C$ channels. $F_r^f, F_r^b$~$\in$~$\mathbb{R}^{H\times W \times C}$ are features for the forward reference frame $r_f$ and backward reference frame $r_b$ respectively.
%
%To capture both low-level and high-level information of an image, we extract the feature maps $F_x$, $F_r^f$ and $F_r^b$ from layers of $relu2\_2$, $relu3\_2$, $relu4\_2$, $relu5\_2$ for $x_t$, $r^{f}$ and $r^{b}$.
%
%Then we find the dense correspondence by calculating the pairwise similarities among these features.
%
We estimate the correspondence matrix by:
\begin{equation}
    \small
	\begin{split}
	\mathbf{C}_t^f&= \textrm{softmax}\left(\frac{\mathbf{F}_x^t {\mathbf{F}_r^f}^{\top}}{\Vert \mathbf{F}_x^t \Vert_2 \Vert \mathbf{F}_r^f \Vert_2}\right), \quad\mathbf{C}_t^f\in \mathbb{R}^{HW \times HW}\\
	\mathbf{C}_t^b &= \textrm{softmax}\left(\frac{\mathbf{F}_x^t {\mathbf{F}_r^b}^{\top}}{\Vert \mathbf{\mathbf{F}}_x^t \Vert_2 \Vert \mathbf{F}_r^b \Vert_2}\right), \quad\mathbf{C}_t^b\in \mathbb{R}^{HW \times HW}
	\end{split}
	\label{eq:correspondence}
\end{equation}
where $\textrm{softmax}(\cdot)$ denotes the softmax operation that is applied to each row of the matrix.
We then generate the warped results of reference images $r^f$ and $r^b$ by:
\begin{equation}
	\begin{split}
	\mathbf{w}_t^f &= \mathbf{C}_t^f\mathbf{r}^f,\\
	\mathbf{w}_t^b &= \mathbf{C}_t^b\mathbf{r}^b,\\
	\end{split}
	\label{eq:warped}
\end{equation}
where $\mathbf{r}^f$ and $\mathbf{r}^b$ denote the vector forms of the reference images $r^f$ and $r^b$.
%We use ``softmax" operation to normalize the $i$-th row the element at $F_x^t$ and all elements at $F_r^f$, $F_r^b$. We reshape the reference images $r^f$, $r^b$ into vectors $v^f$, $v^b$~$\in$~$\mathbb{R}^{HW \times 3}$ . Then, $w_t^f$ and $w_t^b$ can be formulated as follows:
% We formulate $v_t^f$ and $v_t^b$ as follows:
%\begin{equation}
%	\begin{split}
%	w_t^f(i) &= \sum\limits_{j} \mathop{softmax(\mathcal{C}^f(i,j))}\limits_j \odot v^f(j),\\
%	w_t^b(i) &= \sum\limits_{j} \mathop{softmax(\mathcal{C}^b(i,j))}\limits_j \odot v^f(j).
%	\end{split}
%	\label{eq:warped}
%\end{equation}
% where $\odot$ denotes element-wise multiplication.
By applying a reshaping function to $\mathbf{w}_t^f$ and $\mathbf{w}_t^b$, we can get the warped color frames $w_t^f\in\mathbb{R}^{H\times W \times 3}$ and $w_t^b\in\mathbb{R}^{H\times W \times 3}$.
Taking these two warped frames as the input, the bidirectional temporal fusion block is developed to reduce influences of the inaccurate colors of $w_t^f$ and $w_t^b$ and alleviate ghost artifacts or color contamination when large motion happens or severe occlusion occurs (see details in Section~\ref{sec:tfb}).

%\vspace{-2mm}
We note that there usually exist color-bleeding artifacts around the boundaries
of the important objects in videos. To overcome this problem and improve the quality of the colorized videos, we develop a color mapping subnet that contains a mixed expert block (MEB) and a multi-scale recurrent block (MSRB), see details in Section~\ref{sec:meb} and Section~\ref{sec:srb}. This subnet obtains well-colorized image $z_t$ by receiving two inputs: the grayscale input $x_t$ and the fused color map $p_t$ by the bidirectional temporal fusion block.
%
% the semantic segmentation mask $m_t^{s}$ and the edge detection mask $m_t^{e}$.
%
In the color mapping subnet, we first utilize MEB to extract the image prior from the given grayscale input $x_t$. MEB contains a semantic segmentation module that outputs segmentation mask $m_t^{s}$ and an edge detection module that outputs edge detection mask $m_t^{e}$.
We then propose MSRB to further enhance visual performance. Our MSRB is a three-level network based on Unet to obtain a well-colorized image $z_t$. It receives four inputs: the grayscale input $x_t$, the fused color map $p_t$ by the bidirectional temporal fusion block, the segmentation mask $m_t^{s}$ and the edge detection mask $m_t^{e}$.
%
% At the front of the network locates the coarsest level network with 1/4 resolution as inputs. Finer level networks basically have the same structure as in the coarsest level network except that the first convolution layer takes the colorized output from the previous stage as well as its own input features, in a concatenated form. Except for the last finest scale, there is an upconvolution layer before the next stage. At the finest scale, the original resolution color image is restored.
%
The pipeline for \xnet is shown in Figure~\ref{fig:framework}.

% To alleviate color spilling or color bleeding, we introduce two expert modules: (a) semantic segmentation module (b) edge detection module. We borrow a state-of-art semantic segmentation network from ~\cite{protoseg_Siggraph19} as an offline inference module to provide rough semantic mask. We borrow a state-of-art edge detection network from ~\cite{hed_Siggraph19} as an offline inference module to provide trend of pixel changes. These two modules are included to provide more information that make our colorization network easier to distinguish different objects with different colors. Since we already have a coarse colorized frame collected with its semantic information and edge information, what we need to do is just concatenate them together and put into the color mapping subnet. Different from~\cite{Deepexamplar_Siggraph19}, we propose a coarse-to-fine frame, specifically, a three level Unet like structure,  to gain better color mapping performance.
%\vspace{-2mm}
\subsection{Bidirectional Temporal Fusion Block}\label{sec:tfb}
%\vspace{-2mm}
% 目的 时序特征融合， 是用来融合时序特征。 所以我们提出了什么方法，具体怎么做。 现在存在什么问题， 时序的challenage.   Motivated by ,  我们怎么做
% Inspired by \cite{Jiang_2018_CVPR},
The quality of the colorized video frames often decays quickly when the future frames do not match the semantic contents of reference frames.
This makes it a quite common scenario in exemplar-based video colorization tasks that the semantic correspondence is wrongly matched. Hence, it is necessary to utilize temporal clues to select well-colorized regions.
%
% This makes mismatched semantic correspondence a quite common scenario in video colorization task. Hence, it is necessary to leverage temporal clues to select well-colorized regions.
%
% In the correspondence subnet, it’s obvious that the closer the input frame is to the reference frames, the semantic content of these frames can be better matched.(420)
% For relatively long video clips, the semantic content of the two reference images is different, resulting in each reference image providing different semantic content.（429，或删除）
% Therefore, a well-designed fusion mask M will significantly enhance the temporal consistency. Our scheme for designing the fusion mask M is to give more weights to Wft for input frames closer to rf and more weights to Wbt for input frames closer to rb.（431）
% In order to further obtain a more appropriate fusion mask M：（440）

%
To this end, we develop a bidirectional temporal fusion block to obtain the fused color map $p_t$ at time $t$ based on the estimated warped color images $w_{t}^{f}$ and $w_{t}^{b}$ as detailed above.
The fusion of $w_{t}^{f}$ and $w_{t}^{b}$ is mainly based on the temporal distance between the input frame $x_t$ and the reference images $r_f$ and $r_b$.
The temporal distance $d_t$ from input frame $x_t$~$\in$~$\{x_0, x_1,...,x_{N-1} \}$ to forward reference image $r^f$ is defined as $d_t^f = t - 0$ and the temporal distance from input frame $x_t$ to backward reference image $r^b$ is defined as $d_t^b = N-1-t$.
We normalize these temporal distances based on the maximum distance $N-1$ between the input frames, $\tilde{d}_t^f = \frac{t - 0}{N-1}$ and $\tilde{d}_t^b = \frac{N-1-t}{N-1}$.
Then we utilize $\tilde{d}_t^f$ and $\tilde{d}_t^b$ as weight parameters to obtain the fused color frame $p_t$.
When the input frame $x_t$ is closer to $r^f$($r^b$), the corresponding weights for $w^f_t$($w^b_t$) will be larger so that a more accurate fused color map $p_t$ can be obtained.
%
% If the input frame $x_t$ is closer to $r^{f}$, then $w^{f}_t$ increases. Similarly, $w^{f}_t$ increases when $x_t$ is closer to $r^{b}$
% We give more weights to $W^{f}_t$ for input frames closer to $r^{f}$ and more weights to $W^{b}_t$ for input frames closer to $r^{b}$.
%
%
% Then, the output of temporal fusion block can be formulated as below:
% \begin{equation}
%     p_t = \lambda_f W^{f}_t + \lambda_b W^{b}_t,
%     \label{eq:tmb1}
% \end{equation}
% % $\lambda$ is a weight parameter
% where $\lambda_f$ and $\lambda_b$ are weight parameters. We make sure that $\lambda_f + \lambda_b  = 1$.
% \textcolor[rgb]{1.00,0.00,0.00}{To gain the fusion mask $M$},
% In order to further obtain a more appropriate fusion mask $M$,
% Based on such consideration, we use weights that are proportional to the temporal distance between current input frame $x_{t}$ and the reference images $r^{f}$ or $r^{b}$. The weight parameters are formulated as:
%
Based on such consideration, we can obtain the fused color map $p_t$ by:
%
% \begin{equation}
%     \lambda_f = \tilde{d}_t^f = \frac{t - 0}{N - 1},  \lambda_b = \tilde{d}_t^b =\frac{N -1 - t}{N - 1}.
%     \label{eq:tmb2}
% \end{equation}
% Finally, combining Eq.~\ref{eq:tmb1} and Eq.~\ref{eq:tmb2}, we can synthesize the fused color map $p_t$ as follows:

\begin{equation}
      p_t = \frac{t - 0}{N - 1} w^{f}_t + \frac{N-1-t}{N - 1}  w^{b}_t.
    \label{eq:tmb3}
\end{equation}

\subsection{Mixed Expert Block}\label{sec:meb}
Existing methods based on deep neural networks for video colorization often suffer from the color-bleeding artifact, a problematic color spreading near the boundaries between adjacent objects.
Such color-bleeding artifacts affect the quality of generated videos, limiting the applicability of colorization models in practice.
%
%We note that the color bleeding or color spilling usually happens due to lack of segmentation information.
%
Therefore, it is necessary to explore prior knowledge from frames, which can better model the boundaries between adjacent objects to guide colorization.
%
% Given a grayscale input image, out goal is to predict its color channel. A straightforward way is to accomplish this is to train a end-to-end neural network \cite{cic} to directly output the ab channel of image $L$. Currently, most methods like \cite{zhang2019deep} consider neural network as a color mapping function. For example, given a feature map, the neural network is trained to mapping each pixel a target color. However, these methods usually suffer from color bleeding or color spilling. To handle this problem,
% following \cite{zhou2022rethinking} seg
% following \cite{Xie_2015_ICCV} edge

% \textcolor[rgb]{1.00,0.00,0.00}{We develop a mixed expert block into video colorization tasks.????????}
%
In this paper, we develop a mixed expert block (MEB) to model the boundaries between adjacent objects.
Our MEB contains two models, a semantic segmentation model and an edge detection model. MEB takes the grayscale frame $x_t$ as input and outputs two masks, a probability mask of semantic segmentation (see $m_t^{s}$ in Figure~\ref{fig:framework}) and an edge detection mask (see $m_t^{e}$ in Figure~\ref{fig:framework}).
To enhance visual performance near the object boundaries, we use an off-line pre-trained semantic segmentation model~\cite{zhou2022rethinking} that takes the grayscale frame $x_t$ as input to obtain $m_t^{s}$~$\in$~$\mathbb{R}^{H \times W \times C_{Seg}}$, where $C_{Seg}$ is the number of semantic labels.
%
% Firstly, we use an off-line semantic segmentation model to obtain a segmentation mask and utilize this mask to enhance visual performance near object boundaries.
%
% First, we use an off-line semantic segmentation model to obtain a segmentation mask that provides \textcolor[rgb]{1.00,0.00,0.00}{significant information????what does significiant informtion mean???} when rendering colors near object boundaries.
%
In addition, we use an off-line pre-trained edge detection model~\cite{Xie_2015_ICCV} to obtain $m_t^{e}$~$\in$~$\mathbb{R}^{H \times W \times 1}$.
We use the obtained $m_t^{s}$ and $m_t^{e}$ to better guide the colorization.
%$m_t^{e}$ leverages significant spatial clues for the scale-recurrent block in Section~\ref{sec:srb}.
%
% Secondly, to generate more natural and reasonable color images, we introduce an off-line edge detection model to acquire edge mask.
%
% This edge mask leverages spatial clues on refinement of the fused color map $p_t$.
%
% The segmentation mask contains 21 channels, each representing the possibility of one semantic label.
%
% The edge mask tells us the strength of object edges.
%
% Finally, we concatenate these masks together in channel dimension and output the concatenated features as part of input of the following color mapping subnet.

% \textcolor[rgb]{1.00,0.00,0.00}{The edge mask is a single-channel map ranges from zero to one, representing the strength of object edges???? WHY SINGLE HERE!!!!!}. \textcolor[rgb]{1.00,0.00,0.00}{We borrow semantic segmentation model from~\cite{zhou2022rethinking} and edge detection model from~\cite{Xie_2015_ICCV}.????? So What????? what's your goal?}

\subsection{Multi-Scale Recurrent Block}\label{sec:srb}
%
%To generate better colorized videos, result both being global realistic, and preserving sharp color details at object boundaries, an effective solution is to put the colorization network under a coarse-to-fine framework.
To further improve the quality of the colorized videos, we employ a multi-scale recurrent block (MSRB) based on the commonly used coarse-to-fine strategy~\cite{Pan_2020_CVPR, gopro2017}.
%
%We develop a scale-recurrent block (MSRB) that aims at producing vivid and well-structured video frames.
%
MSRB contains three levels, each level is built by the same Unet structure (denoted as $\mathcal{N}$, $\mathcal{N}_{2}$, $\mathcal{N}_{3}$).
It takes the concatenation results of the grayscale input $x_t$, the fused color map $p_t$, the semantic segmentation mask $m_t^{s}$ and the edge detection mask $m_t^{e}$ as the input.
%
%We concatenate those features together in channel dimension and denote the combined feature as $I$.
%
% Then we build SRB each level by a Unet structure.
%
% Firstly,
% In our Scale-Recurrent Block, we build each level by a Unet structure.
%
Specifically, let $I_t$ denote the input of MSRB, we first downsample $I_t$ to the 1/2 resolution feature $I_t^{\frac{1}{2}}$ and a 1/4 resolution feature $I_t^{\frac{1}{4}}$.
Second, we apply a $\mathcal{N}_{3}$ to the feature $I_t^{\frac{1}{4}}$ and obtain a coarse colorized frame $z_t^{\frac{1}{4}}$. In this level, our target is to train  $\mathcal{N}_{3}$ good at controlling the global color style.
For the second level, the network $\mathcal{N}_{2}$ takes the concatenation of $I_t^{\frac{1}{2}}$ and upsampled colorized frame $z_t^{\frac{1}{4}}$ as input and generates the finer colorized frame $z_t^{\frac{1}{2}}$.
Similarly, the network $\mathcal{N}$ takes the concatenation of the original resolution feature $I_t$ with the upsampled result of $z_t^{\frac{1}{2}}$ as input and restores the colorized frame $z_t$ at the original resolution.
%
% \textcolor[rgb]{1.00,0.00,0.00}{At the front of the network locates the coarsest level network and 1/4 resolution intermediate features $I_1$ as input. Then  the finer level network takes 1/2 resolution intermediate features $I_2$ as input.???? Waht do you want to do??????????????????}
% %
% Finer level networks basically have the same structure as in the coarsest level network except that the first convolution layer takes the colorized output from the previous stage as well as its own input features, in a concatenated form.
% %
% Except for the last finest scale, there is an upsampling layer before the next stage. At the finest scale, the original resolution color image $C_3$ restored.
% %
% \textcolor[rgb]{1.00,0.00,0.00}{The network details are shown in Figure~\ref{} ADD Corresponding information here!!!!}

% \begin{figure}[!t]
%     \centering
%     \includegraphics[width=\linewidth,scale=1.00]{images/Coarse-to-FIne_1101.png}\vspace{-0.6em}
%     \caption{Architecture of coarse-to-fine frame. \textcolor[rgb]{1.00,0.00,0.00}{THis FIGURE does not contain any useful information!!!!}
%     }
%     \label{fig:coarse-to-fine}
%     \vspace{-1em}
% \end{figure}

\subsection{Loss Function}
%The goal of our network is to generate temporal coherent and visual-pleasant videos given consecutive video frames. Besides, the style of colorized videos should be faithful to the reference image. Zhang et al.\cite{zhang2019deep} impose the following losses consist of perceptual loss, contextual loss, smoothness loss, adversarial Loss, temporal consistency loss and l1 loss, to achieve realistic video colorization without temporal flickering.
% Combining all the above losses, and the overall objective is:
% 		\begin{equation}
% 			\begin{split}
% 			\mathcal{L}_{I} = & \lambda_{perc}\mathcal{L}_{perc} + \lambda_{context}\mathcal{L}_{context} +
% 			\lambda_{smooth}\mathcal{L}_{smooth}\\&+ \lambda_{adv}\mathcal{L}_{adv}
% 			+ \lambda_{temporal}\mathcal{L}_{temporal}
% 		    + \lambda_{L1}\mathcal{L}_{L1}
% 			\end{split}
% 		\end{equation}
% 		where $\lambda$ controls the relative importance of terms.

To better constrain the network training, we use the edge-enhancing loss by~\cite{Kim_2021_ICCV} to alleviate the color-bleeding artifacts and the problematic color spreading near the boundaries between adjacent objects:
\begin{equation}
	\mathcal{L}_{edge} = \Vert \mathcal{S}(x_t) - \mathcal{S}(z_{t})\Vert_2,
	\label{eq:edgeloss}
\end{equation}%\vspace{-1.5em}
where $\mathcal{S}$ is the Sobel filter used in~\cite{Kim_2021_ICCV}.
%Given an image $x$, $\mathcal{S}(x)$ is defined as:
%\addtolength{\jot}{2pt}
%\begin{equation}
  %\begin{split}
  %&\mathcal{S}(x)= \sqrt{(G_{h} \otimes x)^2 + (G_{v} \otimes x)^2},\\
 %   &G_h= \begin{pmatrix}
%        1 & 0 & \matminus1 \\
%        2 & 0 & \matminus2 \\
%        1 & 0 & \matminus1
%        \end{pmatrix},\hspace{0.1cm}
%   G_v= \begin{pmatrix}
%        1 & 2 & 1 \\
%        0 & 0 & 0 \\
%        \matminus1 & \matminus2 & \matminus1
%       \end{pmatrix},\\
%\end{split}
%\label{eq:eq1}
%\end{equation}
%
%where $G_h$ and $G_v$ are horizontal and vertical filter kernels, and $\otimes$ denotes the convolution operation.
%

We note that the majority of the colorized areas are smooth, artifacts usually happen at object boundaries. Simply using the $L_1$-norm-based content aware loss function (i.e., $\Vert y_t - z_{t}\Vert_1$, where $z_{t}$ denotes the colorized frame produced by our \xnet and $y_t$ denotes the ground truth frame.) would make the trained colorization model mainly focus on well-colorized regions and ignore the regions where artifacts exist.
To overcome this problem, we further employ the hard example mining loss  by~\cite{Pan_2020_CVPR, Xu_2019_CVPR}.
%
% \textcolor[rgb]{1.00,0.00,0.00}{To handle this issue, we propose to focus on the difficult areas by means of sorting all pixels in a descending order of the $L_1$-norm-based content aware loss to encourage the model be aware of hard examples.}
%
%We represent $L_1$-norm-based content aware loss as:
% \begin{equation}
%	\mathcal{L}_{content} = \Vert y_t - z_{t}\Vert_1,
%	\label{eq:l1}
%\end{equation}
%
% \textcolor[rgb]{1.00,0.00,0.00}{where $\tilde{x}_{t}^{ab}$ denotes the colorized frame at time $t$ $ab$,  denotes the chrominance in CIE $Lab$ color space.  the generated and $x_t^{ab}$ denotes the }
%where $z_{t}$ denotes the colorized frame produced by our \xnet and $y_t$ denotes the ground truth frame.
%
%$ab$ denotes the chrominance in CIE $Lab$ color space.
%
%Combining $L_1$-norm-based content aware loss function~\eqref{eq:l1}, the hard example mining loss is defined as:
% \vspace{-5pt}
%\begin{equation}
%\mathcal{L}_{hem} = \mathcal{L}_{content} + \lambda \Vert M^h \odot(y_t - %z_{t})\Vert_1,
%\end{equation}
% \vspace{-1.5em}
%where $M^h$ is the binary mask indicating the hard regions~\cite{}, $\lambda$ is a weight parameter, and $\odot$ denotes element-wise multiplication.
% \begin{equation}
% \mathcal{L}_{hem} = \Vert S(I^l) - S(I^{ab})\Vert_2
% \label{eq:hemloss}
% \end{equation}\vspace{-1.5em}

In addition to those above loss functions, we further use the same loss function based on the content, perception and temporal consistency by~\cite{zhang2019deep} to constrain the network training. Finally, the loss function for the network training is:
\begin{equation}
\begin{split}
\mathcal{L}_{total} =  \lambda_{edge}\mathcal{L}_{edge} + \lambda_{hem}\mathcal{L}_{hem} + \lambda_{c}\mathcal{L}_{c},
\end{split}
\end{equation}
where $\mathcal{L}_{c}$ is the loss function by~\cite{zhang2019deep}; $\mathcal{L}_{hem}$ is the hard example mining loss function by~\cite{Pan_2020_CVPR}; $\lambda_{edge}$, $\lambda_{hem}$, and $\lambda_{c}$ are weight parameters.

% 1107 breakdance
\begin{figure*}[!t]\footnotesize
 \begin{center}
  \begin{tabular}{cccccccc}
  \multicolumn{3}{c}{\multirow{5}*[60pt]{\includegraphics[width=0.368\linewidth, height=0.295\linewidth]{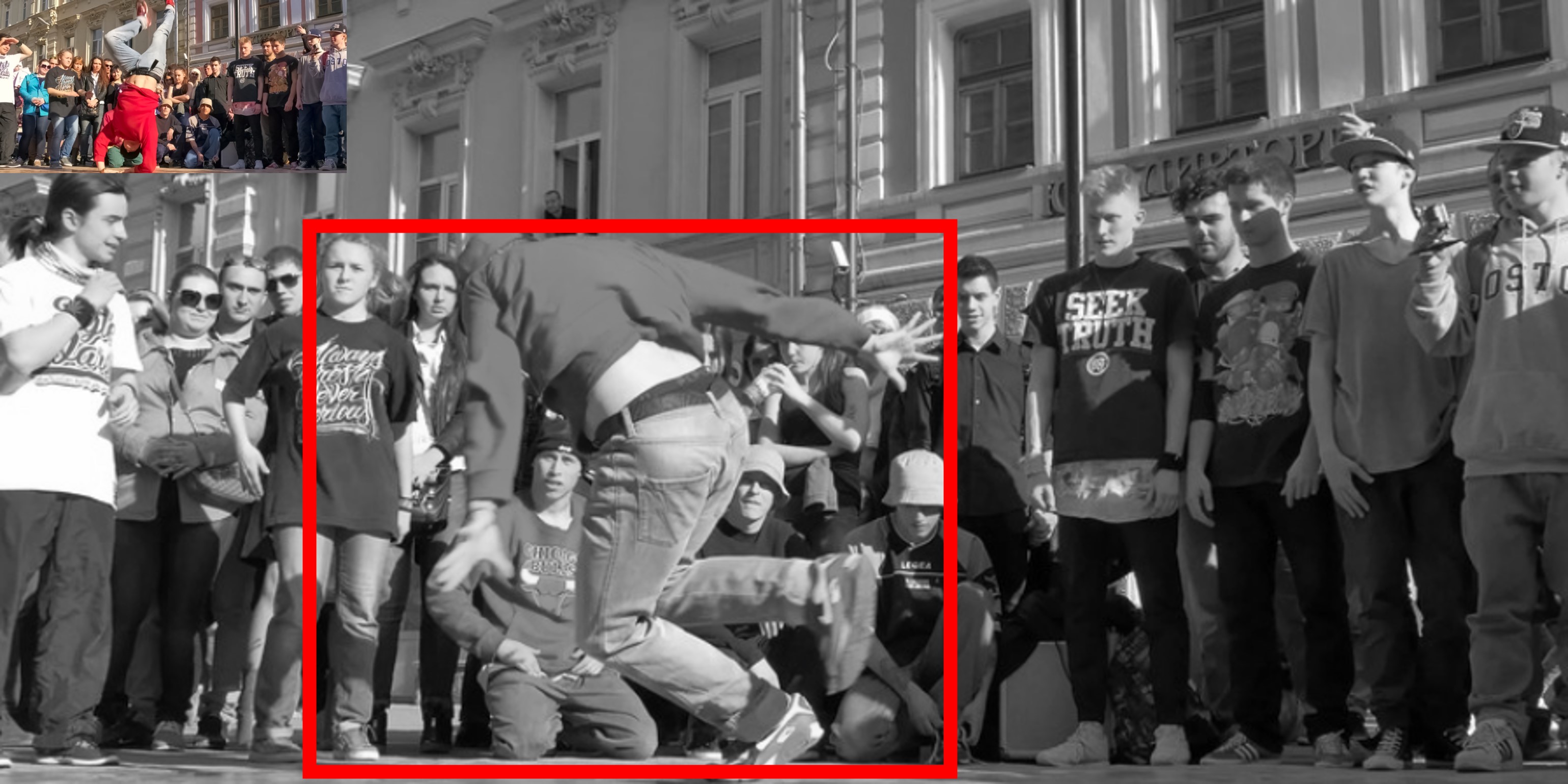}}}&\hspace{-3.5mm}
  \includegraphics[width=0.15\linewidth, height = 0.132\linewidth]{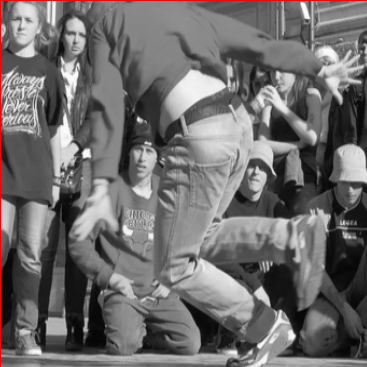} &\hspace{-3.5mm}
  \includegraphics[width=0.15\linewidth, height = 0.132\linewidth]{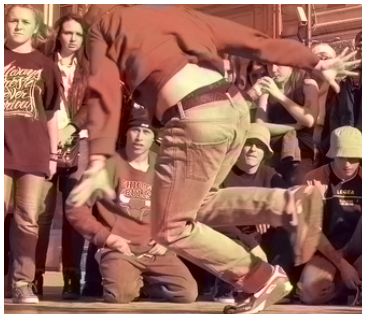} &\hspace{-3.5mm}
  \includegraphics[width=0.15\linewidth, height = 0.132\linewidth]{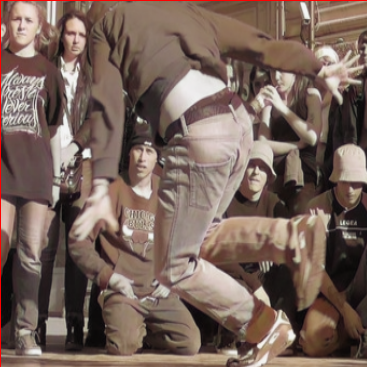} &\hspace{-3.5mm}
  \includegraphics[width=0.15\linewidth, height = 0.132\linewidth]{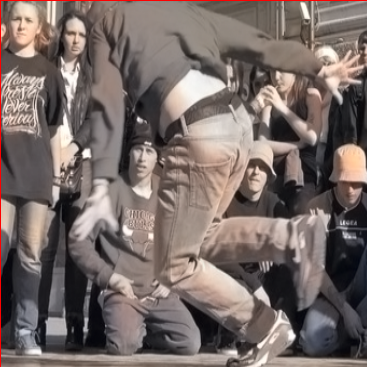} \\
  \multicolumn{3}{c}{~} &\hspace{-3.5mm}  (a) Input &\hspace{-3.5mm}  (b) CIC~\cite{cic} &\hspace{-3.5mm}  (c) FAVC~\cite{favc}  &\hspace{-3.5mm}  (d) DeepRemaster~\cite{IizukaSIGGRAPHASIA2019}\\

  \multicolumn{3}{c}{~} & \hspace{-3.5mm}
  \includegraphics[width=0.15\linewidth, height = 0.132\linewidth]{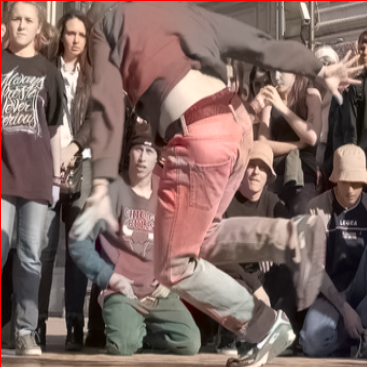} & \hspace{-3.5mm}
  \includegraphics[width=0.15\linewidth, height = 0.132\linewidth]{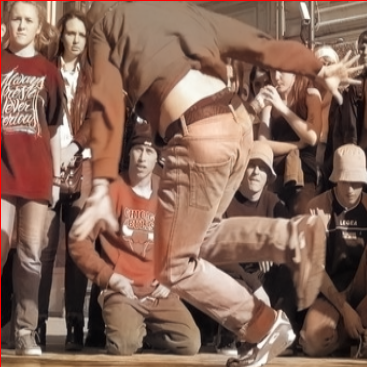} & \hspace{-3.5mm}
\includegraphics[width=0.15\linewidth, height = 0.132\linewidth]{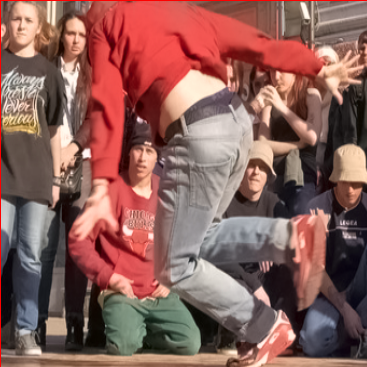} & \hspace{-3.5mm}
  \includegraphics[width=0.15\linewidth, height = 0.132\linewidth]{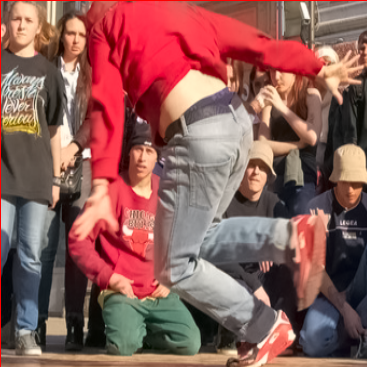} \\
  \multicolumn{3}{c}{\hspace{-3.5mm} Input frame and the reference image } &\hspace{-3.5mm}  (e) DeepExemplar~\cite{zhang2019deep} &  \hspace{-3.5mm} (f) TCVC~\cite{liu2021temporally}  &\hspace{-3.5mm}  (g) Ours  & \hspace{-3.5mm}(h) Ground Truth \\
 \end{tabular}
 \end{center}
 \vspace{-5mm}
 \caption{Qualitative colorization comparisons on clip \textit{breakdance} from the DAVIS dataset~\cite{Perazzi_CVPR_2016}. The evaluated methods do not generate colorful frames, and the colors by these methods are not estimated correctly in (b)-(f). Our method generates well-colorized image.}
 \label{fig:breakdance}
\end{figure*}
% \vspace{-2mm}
% 1111b
\begin{figure*}[!t]\footnotesize
 \begin{center}
  \begin{tabular}{cccccccc}
  \multicolumn{3}{c}{\multirow{5}*[60pt]{\includegraphics[width=0.368\linewidth, height=0.295\linewidth]{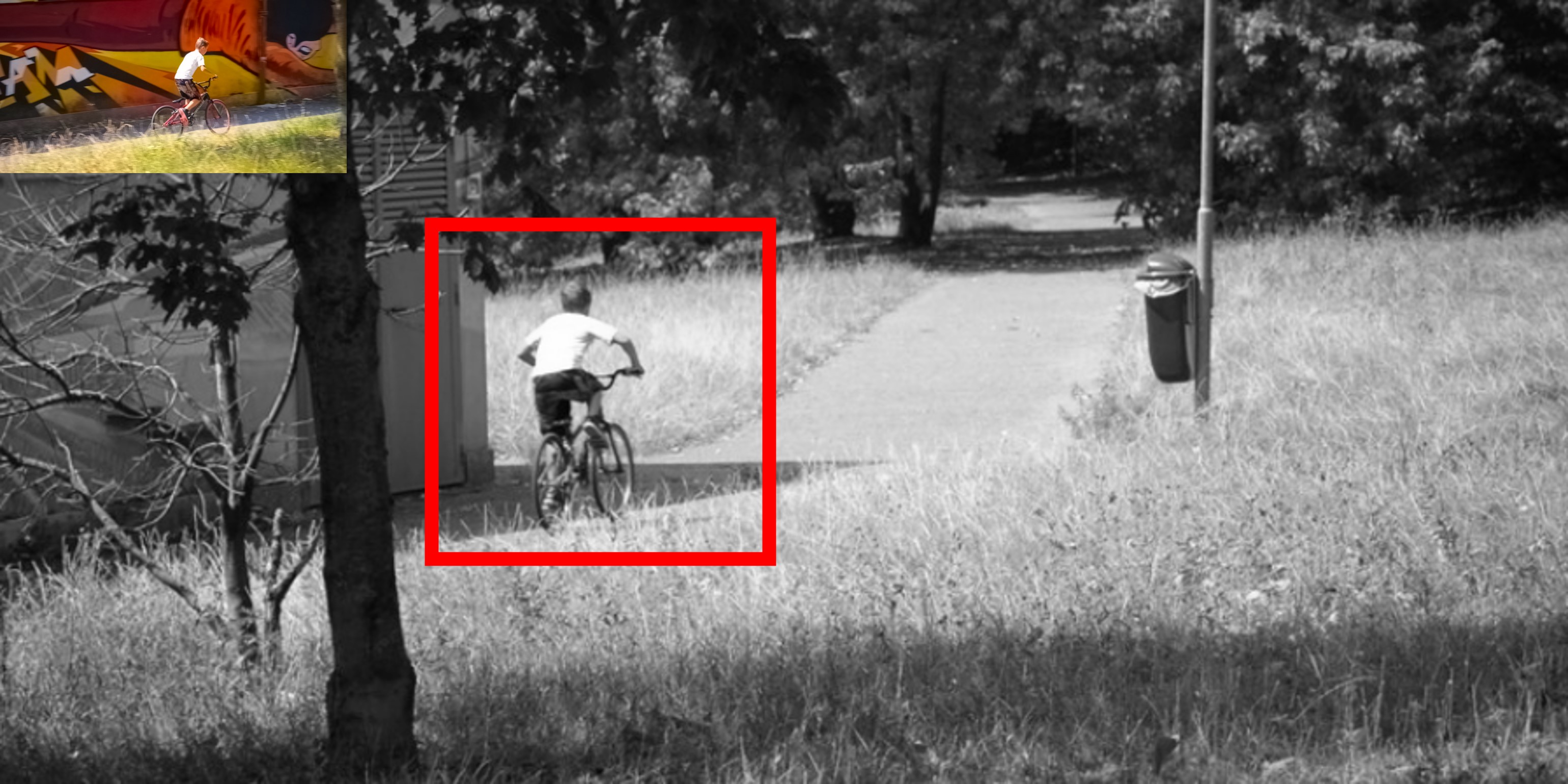}}}&\hspace{-3.5mm}
  \includegraphics[width=0.15\linewidth, height = 0.132\linewidth]{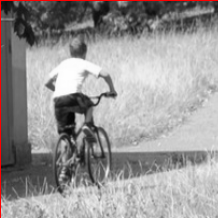} &\hspace{-3.5mm}
  \includegraphics[width=0.15\linewidth, height = 0.132\linewidth]{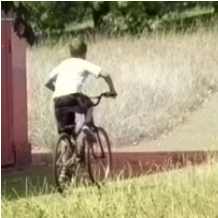} &\hspace{-3.5mm}
  \includegraphics[width=0.15\linewidth, height = 0.132\linewidth]{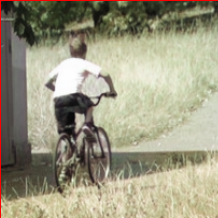} &\hspace{-3.5mm}
  \includegraphics[width=0.15\linewidth, height = 0.132\linewidth]{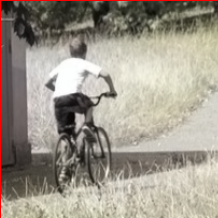} \\
  \multicolumn{3}{c}{~} &\hspace{-3.5mm}  (a) Input &\hspace{-3.5mm}  (b) CIC~\cite{cic} &\hspace{-3.5mm}  (c) FAVC~\cite{favc}  &\hspace{-3.5mm}  (d) DeepRemaster~\cite{IizukaSIGGRAPHASIA2019}\\

  \multicolumn{3}{c}{~} & \hspace{-3.5mm}
  \includegraphics[width=0.15\linewidth, height = 0.132\linewidth]{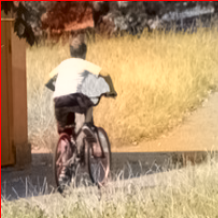} & \hspace{-3.5mm}
  \includegraphics[width=0.15\linewidth, height = 0.132\linewidth]{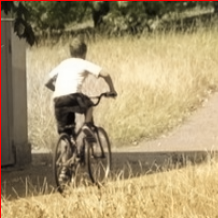} & \hspace{-3.5mm}
  \includegraphics[width=0.15\linewidth, height = 0.132\linewidth]{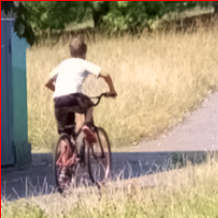} & \hspace{-3.5mm}
  \includegraphics[width=0.15\linewidth, height = 0.132\linewidth]{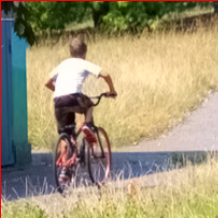} \\
  \multicolumn{3}{c}{\hspace{-3.5mm} Input frame and the reference image} &\hspace{-3.5mm}  (e) DeepExemplar~\cite{zhang2019deep} &  \hspace{-3.5mm} (f) TCVC~\cite{liu2021temporally}  &\hspace{-3.5mm}  (g) Ours  & \hspace{-3.5mm}(h) Ground Truth \\
 \end{tabular}
 \end{center}
 \vspace{-5mm}
 \caption{{Qualitative colorization comparisons on clip \textit{bmx-trees} from the DAVIS dataset~\cite{Perazzi_CVPR_2016}. Our method in (g) is able to avoid color contamination near the boundaries of lawns and roads.}}
 \label{fig:bmx-trees}
\end{figure*}

\section{Experimental Results}\label{sec:experiments}
In this section, we evaluate the proposed method against state-of-the-art ones. Due to the page limit, we include more results in the supplemental material. The training code and test model will be available to the public.
%
% In order to compare the performance with other state-of-the-art methods, single and dual reference frames are adopted to evaluate our model, respectively.
%
% We evaluate the proposed \xnet on benchmark datasets and \textcolor[rgb]{1.00,0.00,0.00}{experimental settings for video colorization task??? Waht are you talin about? Evaluting setting???????????}.
%
%
% In Section 4.1, we present detailed comparison results of different methods.
%

% \textcolor[rgb]{1.00,0.00,0.00}{To begin with, we evaluate our model with single reference frame to provide fair comparation to recent state-of-art methods. Then we evaluate our model with two reference frame to show the improvement of colorization performance. ??????!!!! }

\noindent\textbf{Datasets.}
% The quantitative results are summarized in Table~\ref{tab:psnr} . Following \cite{liu2021temporally},
We adopt the DAVIS dataset~\cite{Perazzi_CVPR_2016} and the Videvo dataset~\cite{Lai2018} as the benchmark datasets for training and testing.
Our training set includes 60 clips from DAVIS and 80 clips from Videvo, whereas our testing set includes 30 clips from DAVIS.
Before training, we resize each frame of all training videos into 384 $\times$ 224 pixels.

To obtain the reference color images, we adopt a similar strategy in ~\cite{IizukaSIGGRAPHASIA2019}.
We provide the first and the last frames as the reference images for each 90-frame length video clip.
Then, we utilize these reference images to colorize the remaining part (the consecutive video frames from frame 6 to frame 85) of the video clips.
%
% Specifically, we use the consecutive video frames from frame 5 to frame 85 as input frames.
%
% We conduct our experiments mainly on the DAVIS dataset~\cite{Perazzi_CVPR_2016} and the Videvo dataset\cite{videvo, Lai2018}. The test set of the DAVIS dataset consists of 30 video clips of various scenes. There are about 30 to 100 frames in each video clip. The test set of the Videvo dataset contains 20 videos and each one has about 300 video frames. In totally, we evaluate our models and baselines on 50 test videos. All the videos are resized to 480p in both datasets.

\noindent\textbf{Evaluation metrics.}
To evaluate the quality of the colorized videos, we use the peak signal-to-noise ratio (PSNR), the structural similarity index (SSIM)~\cite{wang2004image}, and the learned perceptual image patch similarity (LPIPS)~\cite{zhang2018unreasonable} matrices. The color distribution consistency index (CDC)~\cite{liu2021temporally} is used to measure the temporal consistency of videos.

\noindent\textbf{Implementation details.}
We train all models using the PyTorch framework on a machine with four RTX-A6000 GPUs.
We adopt the CIE \textit{LAB} color space for each frame in our experiments.
In the training process, we use the Adam algorithm~\cite{kingma2014adam} with default parameters and train our models with 50 epochs.
%
% We resize all the training images to 384 × 224.
%
The batch size is set to 8. The learning rate is set to a constant $2 \times {10}^{-4}$. We employ the off-the-shelf method ProtoSeg~\cite{zhou2022rethinking} for semantic segmentation. We also employ HED~\cite{Xie_2015_ICCV} for edge detection with default parameters. For the weights in the loss function, we empirically set $\lambda_{edge}=2$, $\lambda_{hem}=2$, $\lambda_{c}=1$.
% We use training iterations 1000k and batch size equals to 1. Adam algorithm~\cite{kingma2014adam} is adopted with a learning rate of $2\times10^{-3}$, we set $\lambda_a=6.6\times10^{-3}$ and $\lambda_p=1$. WAhT do these mean?????????????????
% We train all models using four RTX-A6000 GPUs. We use the Adam algorithm~\cite{kingma2014adam} with default parameters settings,  \textcolor[rgb]{1.00,0.00,0.00}{$\lambda_a=6.6\times10^{-3}$ and $\lambda_p=1$ . use training iterations 1000k and batch size equals to 1. Adam algorithm~\cite{kingma2014adam} is adopted with a learning rate of $2\times10^{-3}$, we set $\lambda_a=6.6\times10^{-3}$ and $\lambda_p=1$. WAhT do these mean??????????????????????????????}
% %
% The batch size is set to be ?????. The patch size is ??????????.
% %
% The total number iterations are set to be 1000k.
% %
% \textcolor[rgb]{1.00,0.00,0.00}{EXPLAIN OTHER SETTINFS HERE!!!!!}.

% %%%%%%%%%%%%%%%%%%%%%%%%%%%%%%%%%%%%%%%%%%%%%%%
% \input{Figures/deblurring_fig.tex}
% %%%%%%%%%%%%%%%%%%%%%%%%%%%%%%%%%%%%%%%%%%%%%%%

% %%%%%%%%%%%%%%%%%%%%%%%%%%%%%%%%%%%%%%%%%%%%%%
% \input{Tables/dualpixel_deblurring_table}
% \input{Figures/dualpixel_deblurring_fig}
% %%%%%%%%%%%%%%%%%%%%%%%%%%%%%%%%%%%%%%%%%%%%%%%

% %%%%%%%%%%%%%%%%%%%%%%%%%%%%%%%%%%%%%%%%%%%%%%
% \input{Tables/gaussian_denoising_table}
% \input{Tables/real_denoising_table}
% %%%%%%%%%%%%%%%%%%%%%%%%%%%%%%%%%%%%%%%%%%%%%%%
% %%%%%%%%%%%%%%%%%%%%%%%%%%%%%%%%%%%%%%%
% \input{Figures/denoising_fig}
% %%%%%%%%%%%%%%%%%%%%%%%%%%%%%%%%%%%%%%%

% \input{Tables/ablation_table1}
%%%%%%%%%%%%%%%%%%%%%%%%%%%%%%%%%%%%%%%%%%%%%%

\subsection{Comparisons with State-of-the-Art Methods}
We compare the proposed method with state-of-the-art ones including both the automatic colorization ones~\cite{favc, liu2021temporally} and exemplar-based video colorization ones~\cite{IizukaSIGGRAPHASIA2019, zhang2019deep}. In addition, to make the paper more self-contained, we include comparisons with state-of-the-art image colorization methods~\cite{let, cic, tog17, inscolor}.
We note that exemplar-based methods~\cite{zhang2019deep, IizukaSIGGRAPHASIA2019} require a reference exemplar as guidance. For fair comparisons, we choose the first ground truth color frame as the reference exemplar for these methods.

Table~\ref{tab:psnr} shows the quantitative evaluation results on the commonly used benchmark~\cite{Perazzi_CVPR_2016}, where the proposed \xnet generates better-colorized results.
In particular, \xnet outperforms DeepExemplar~\cite{zhang2019deep}, a state-of-the-art exemplar-based method, by up to 1.37 dB in terms of PSNR, and achieves 18.52$\%$ improvement in terms of the temporal consistency index CDC.
Compared to the TCVC method~\cite{liu2021temporally} which mainly focuses on improving the temporal consistency of the colorized videos, our \xnet still shows an improvement of 10.9$\%$ in terms of CDC.
All comparisons in Table~\ref{tab:psnr} demonstrate that the proposed \xnet generates better-colorized frames while the generated videos have better temporal consistency.
\begin{figure}[!t]
    \centering
    \includegraphics[width=0.98\linewidth,scale=1.00]{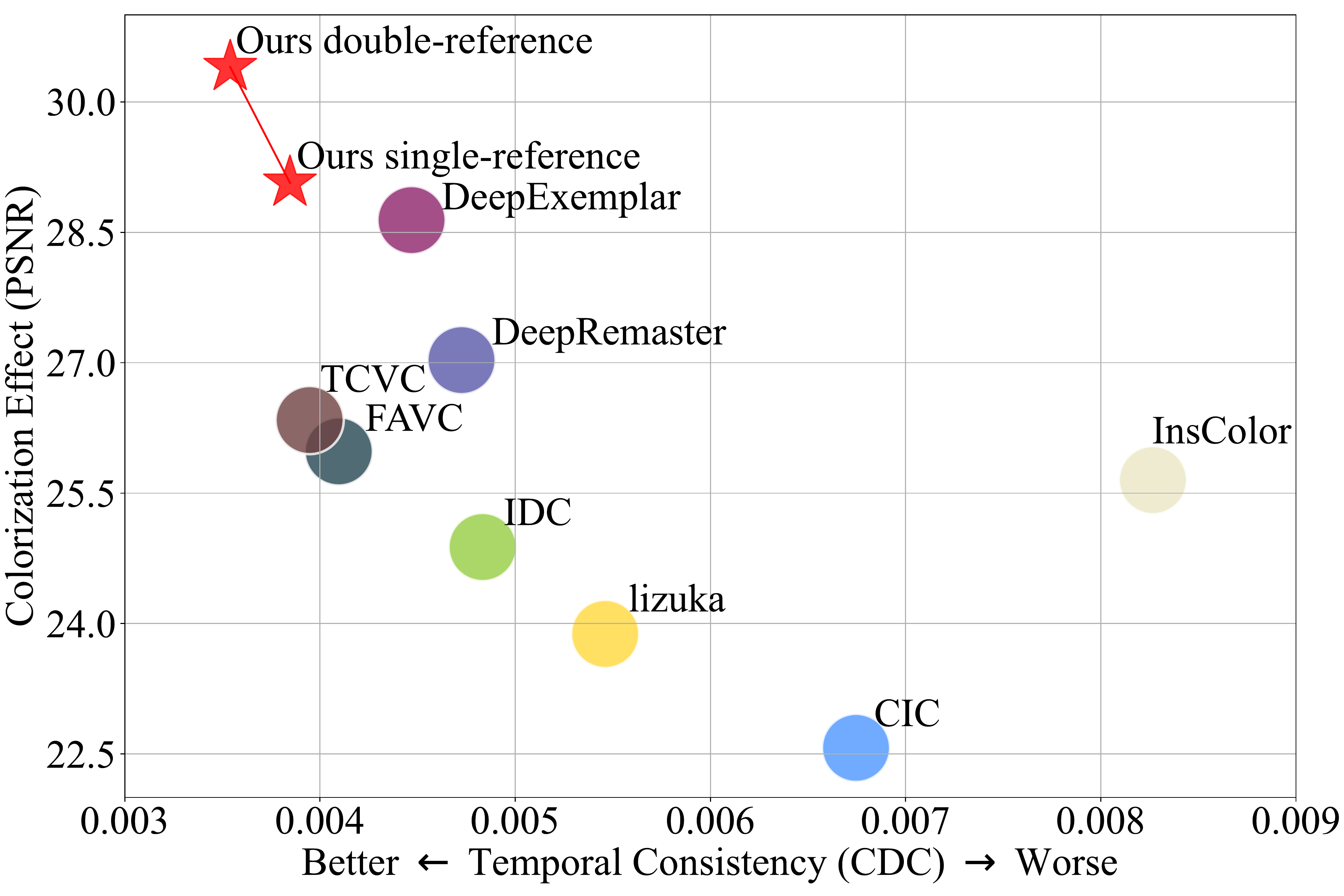}\vspace{-0.6em}
    \caption{Model performance and temporal consistency comparison between our proposed \xnet family and other colorization methods on the DAVIS dataset~\cite{Perazzi_CVPR_2016}. Our \xnet family achieves both a higher PSNR and a lower CDC among all other methods.}
    \label{fig:comp}
    % \vspace{-1em}
    \vspace{-5mm}
\end{figure}
%--------------------------------------------------------

 Figure~\ref{fig:breakdance} and Figure~\ref{fig:bmx-trees} show qualitative comparisons of the proposed \xnet and the methods with top performance in Table~\ref{tab:psnr}.
 We note that the DeepExemplar method~\cite{zhang2019deep} and the DeepRemaster method~\cite{IizukaSIGGRAPHASIA2019} do not generate the results with the correct colors.
 In contrast, the proposed \xnet generates vivid frames whose colors are visually close to the ground truths.
In particular, the proposed \xnet restores the color of the cloth in Figure~\ref{fig:breakdance}.
In addition, the proposed \xnet can avoid color contamination near the boundaries of lawns and roads in Figure~\ref{fig:bmx-trees}, suggesting the effectiveness of the proposed MEB on video colorization tasks.

\subsection{Evaluations on Real-World Grayscale Videos}
\begin{figure*}[!t]\footnotesize
 \begin{center}
  \begin{tabular}{cccccc}
  \multicolumn{3}{c}{\multirow{5}*[60pt]{\includegraphics[width=0.49\linewidth, height=0.295\linewidth]{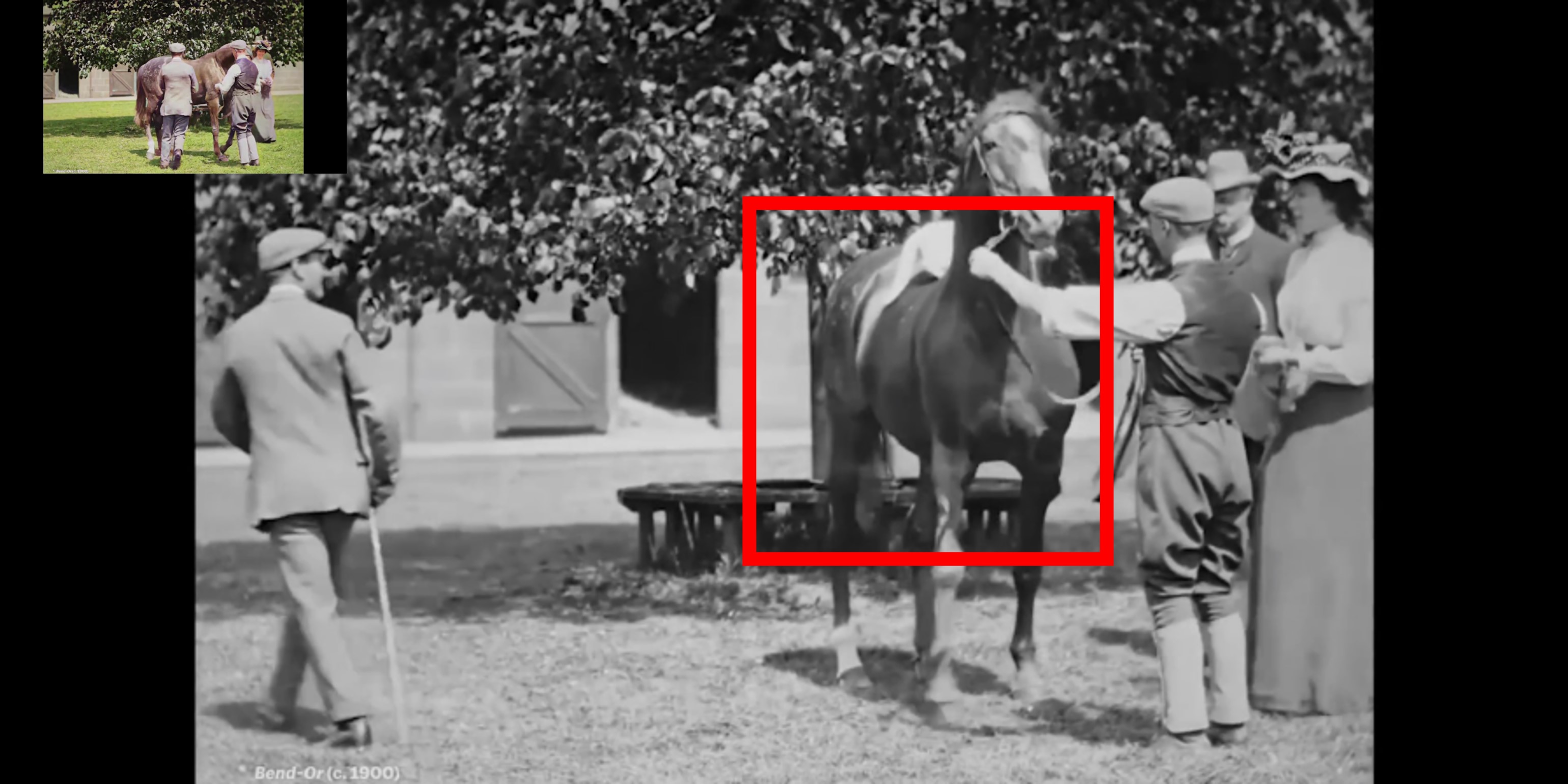}}}&\hspace{-3.5mm}
  \includegraphics[width=0.15\linewidth, height = 0.132\linewidth]{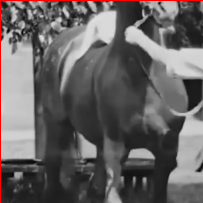} &\hspace{-3.5mm}
  \includegraphics[width=0.15\linewidth, height = 0.132\linewidth]{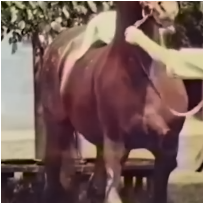} &\hspace{-3.5mm}
  \includegraphics[width=0.15\linewidth, height = 0.132\linewidth]{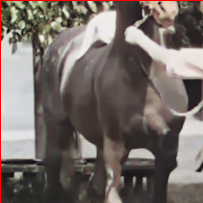} \\
  \multicolumn{3}{c}{~} &\hspace{-3.5mm}  (a) Input &\hspace{-3.5mm}  (b) CIC~\cite{cic}  &\hspace{-3.5mm}  (c) FAVC~\cite{favc}  \\

  \multicolumn{3}{c}{~} & \hspace{-3.5mm}
  \includegraphics[width=0.15\linewidth, height = 0.132\linewidth]{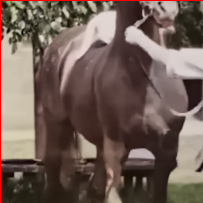} & \hspace{-3.5mm}
  \includegraphics[width=0.15\linewidth, height = 0.132\linewidth]{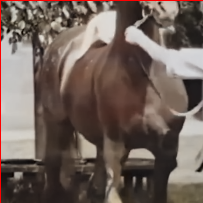} & \hspace{-3.5mm}
  \includegraphics[width=0.15\linewidth, height = 0.132\linewidth]{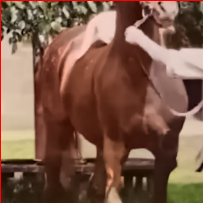} \\
  \multicolumn{3}{c}{\hspace{-3.5mm} Input frame and the reference image} &\hspace{-3.5mm}  (d) DeepExemplar~\cite{zhang2019deep} &  \hspace{-3.5mm} (e) TCVC~\cite{liu2021temporally}  &\hspace{-3.5mm}  (f) Ours   \\
 \end{tabular}
 \end{center}
 \vspace{-6mm}
\caption{{Qualitative colorization comparisons on real-world film \textit{Bend-Or}. The colorization methods~\cite{favc,zhang2019deep,liu2021temporally} do not colorize the horse and the backgrounds well. Our method colorizes the frame well, where the colors look better.}}
\label{fig:bendor}
\end{figure*}

We further evaluate the proposed method on real-world grayscale videos, where the ground truth colors videos are not available.
Figure~\ref{fig:bendor} shows an example from a challenging real-world old film \textit{Bend-Or}.
We select the well-colorized frames from the internet as the reference exemplar for all the exemplar-based methods and compare the proposed method with state-of-the-art ones~\cite{cic, favc, IizukaSIGGRAPHASIA2019, zhang2019deep, liu2021temporally}.
Figure~\ref{fig:bendor} shows that state-of-the-art methods do not colorize the objects (e.g., horse) well.
In contrast, our method generates a better-colorized frame, where the colors of the horse look natural (see Figure~\ref{fig:bendor}(g)).

\vspace{-1mm}
\subsection{Temporal Consistency Evaluations}
Existing methods mainly focus on improving the quality of each colorized frame.
As temporal consistency is one of the important factors that determines the quality of colorized videos, we further demonstrate whether colorized videos by the proposed method have a better temporal consistency or not.
We plot the relationship between the quality of each colorized frame and the temporal consistency in Figure~\ref{fig:comp}.

Overall, the proposed method generates high-quality colorized videos with better temporal consistency.

% 1110 DAVIS new data , without GT
\begin{table}[!t]
	\centering
	\small
 \caption{Quantitative comparisons of the proposed method against state-of-the-art ones on the synthetic DAVIS dataset~\cite{Perazzi_CVPR_2016}. Our method achieves the best performance in terms of PSNR, SSIM, LPIPS, and CDC.}
 \vspace{-2mm}
	\begin{adjustbox}{max width=\linewidth}
		\begin{tabular}{l|cccc}
			
			\noalign{\hrule height 0.3mm}
% 			\rowcolor[HTML]{F5F5F5}
			Method       & PSNR${\uparrow}$ & SSIM${\uparrow}$ & LPIPS${\downarrow}$ & CDC ${\downarrow}$ \\ \hline
			Input             & 19.84           & 0.953      & 0.230              & /           \\
			lizuka et al~\cite{let} & 23.88       & 0.947        & 0.176               & 0.005461           \\
			% SwinIR       & --               & --               & --                  & --               \\
			CIC~\cite{cic}     & 22.57           & 0.947     & 0.106   & 0.006746           \\
			IDC~\cite{tog17}   & 24.88           & 0.949     & 0.116   & 0.004833           \\
			InsColor~\cite{inscolor} & 25.65     & 0.951    & 0.082     & 0.008267           \\
			
			FAVC~\cite{favc}     &    25.98       & 0.967  &    0.172     &    0.004096           \\
			TCVC~\cite{liu2021temporally}  &  26.34   & 0.962  &  0.175  &   0.003947   \\
			
			DeepRemaster~\cite{IizukaSIGGRAPHASIA2019}  & 27.03  & 0.964 &  0.057    &  0.004725           \\
			DeepExemplar~\cite{zhang2019deep}     &    28.64   & 0.972    & 0.041   &   0.004469 \\ \hline
			Ours single-reference &  29.07 & 0.973   & 0.036       & 0.003846      \\
			Ours double-reference & \textbf{30.40} & \textbf{0.976 }  & \textbf{0.030} & \textbf{0.003540}           \\
			\noalign{\hrule height 0.3mm}
			
		\end{tabular}
	\end{adjustbox}
	\label{tab:psnr}
	\vspace{-0.7em}
\end{table}

\begin{table}[!t]
	\small
	\centering
	\setlength{\tabcolsep}{3.8mm}
 \caption{{Quantitative evaluations of each component in the proposed \xnet on the DAVIS dataset~\cite{Perazzi_CVPR_2016}.}}
 \vspace{-2mm}
	\begin{adjustbox}{max width=\linewidth}
		\begin{tabular}{l|cc}
			\noalign{\hrule height 0.3mm}
% 			\rowcolor[HTML]{F5F5F5}
			Method       & PSNR${\uparrow}$ & CDC${\downarrow}$  \\ \hline
			Ours w/o BTFB  & 28.32   & 0.004359   \\
			% SwinIR       & --               & --               & --                  & --               \\
			Ours w/o $m_t^{s}$    & 27.58           & 0.004484             \\
			Ours w/o $m_t^{e}$   & 27.99           & 0.004078            \\
			Ours w/o $m_t^{e}$, $m_t^{s}$   & 26.61           & 0.003972              \\
			Ours w/o MSRB  & 27.29           & 0.004640             \\ \hline
			Ours w/o $\mathcal{L}_{edge}$ & 28.15 & 0.004660    \\
			Ours w/o $\mathcal{L}_{hem}$ & 27.08 & 0.004525    \\
			Ours w/o $\mathcal{L}_{edge}$ \& $\mathcal{L}_{hem}$ & 27.00 & 0.004682    \\
			\hline
% 			\rowcolor[HTML]{F7FAFE}
			Ours        & \textbf{30.40}           & \textbf{0.003540 }                \\
			\noalign{\hrule height 0.3mm}
		\end{tabular}
	\end{adjustbox}
	%\vspace{-0.3em}
	\label{tab:ablation}
	\vspace{-1.0em}
\end{table}

%%%%%%%%%%%%%%%%%%%%%%%%%%%%%%%%%%%%%%%%%%%%%%%
	
% 1107 fanghua ablation
\begin{figure*}[!t]\footnotesize
 \begin{center}
  \begin{tabular}{cccccccc}
  \multicolumn{3}{c}{\multirow{5}*[60pt]{\includegraphics[width=0.368\linewidth, height=0.295\linewidth]{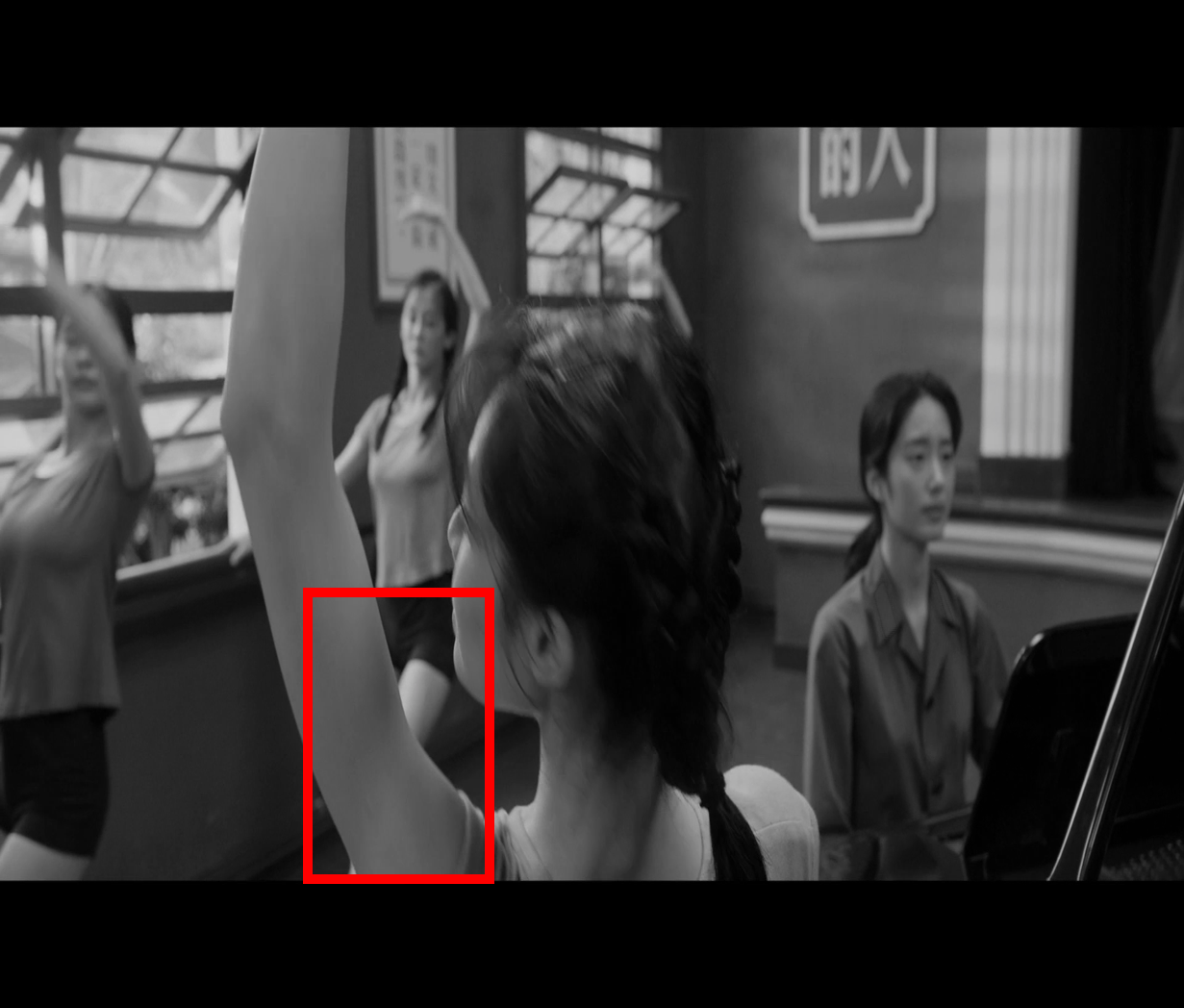}}}&\hspace{-3.5mm}
  \includegraphics[width=0.15\linewidth, height = 0.132\linewidth]{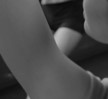} &\hspace{-3.5mm}
  \includegraphics[width=0.15\linewidth, height = 0.132\linewidth]{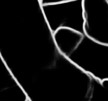} &\hspace{-3.5mm}
  \includegraphics[width=0.15\linewidth, height = 0.132\linewidth]{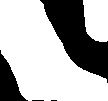} &\hspace{-3.5mm}
  \includegraphics[width=0.15\linewidth, height = 0.132\linewidth]{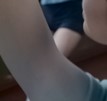} \\
  \multicolumn{3}{c}{~} &\hspace{-3.5mm}  (a) Input &\hspace{-3.5mm}  (b) $m_t^{e}$ &\hspace{-3.5mm}  (c) $m_t^{s}$  &\hspace{-3.5mm}  (d) w/o $m_t^{e}$\&$m_t^{s}$\\

  \multicolumn{3}{c}{~} & \hspace{-3.5mm}
  \includegraphics[width=0.15\linewidth, height = 0.132\linewidth]{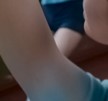} & \hspace{-3.5mm}
  \includegraphics[width=0.15\linewidth, height = 0.132\linewidth]{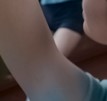} & \hspace{-3.5mm}
  \includegraphics[width=0.15\linewidth, height = 0.132\linewidth]{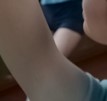} & \hspace{-3.5mm}
  \includegraphics[width=0.15\linewidth, height = 0.132\linewidth]{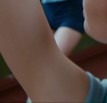} \\
  \multicolumn{3}{c}{\hspace{-3.5mm} Input frame from \textit{Youth}} &\hspace{-3.5mm}  (e) w/o $m_t^{s}$ &  \hspace{-3.5mm} (f) w/o $m_t^{e}$  &\hspace{-3.5mm}  (g) Ours  & \hspace{-3.5mm}(h) Ground Truth \\
 \end{tabular}
 \end{center}
 \vspace{-5mm}
 \caption{Effectiveness of the proposed MEB module. (b) and (c) denote the edge detection mask $m_t^e$ and the semantic segmentation mask $m_t^s$. (d) denotes the colorized image by the proposed model without using both $m_t^e$ and $m_t^s$ (i.e., the proposed method without using the MEB module). (e) denotes the colorized image by the proposed model without using $m_t^{s}$. (f) denotes the colorized image by the proposed model without using $m_t^{e}$. Note that the method without using the MEB module does not colorize the frame well, where the color of the arm is contaminated by the color of the T-shirt. Our model with the MEB module restores better colors in (g).}
 \label{fig:mixed_expert_model}
  \vspace{-2mm}
\end{figure*}

% \section{Ablation Study}
\section{Analysis and Discussion}
In this section, we further conduct extensive ablation studies to demonstrate the effectiveness of the proposed temporal fusion model, mixed expert block, and multi-scale recurrent block.
We train all the possible baseline models using the same settings as the proposed \xnet and evaluate them on the DAVIS dataset~\cite{Perazzi_CVPR_2016} for fair comparisons.
Table~\ref{tab:ablation} shows the quantitative evaluation results.
%
% We test the models on DAVIS dataset.

\noindent \textbf{Effectiveness of the bidirectional temporal fusion block.}
% ！！！
% 是干什么的，作用
% 我们是怎么验证的，  To exam ...    具体的操作， 训练一个baseline
% 从结果来分析
The proposed BTFB is used to utilize temporal clues to select well-colorized regions from the reference exemplars.
%
% It weights pixels transferred from reference images to achieve frame-specific feature fusion.
%
%It gives weights to the bidirectional warped color images based on their temporal distances to the reference images.
%
To demonstrate its effectiveness, we replace the BTFB with average weighting to fuse the bidirectional features. Table~\ref{tab:ablation} shows that our model without TFB does not perform well.
%
% We first consider removing this block. We disable temporal fusion block and let the semantic correspondence passing through color mapping subnet directly. Table~\ref{tab:ablation} shows that the temporal consistency vastly downgrades on the CDC metric, which demonstrates the importance of temporal fusion block (TFB) in \xnet to properly fusing warped color sourced from two reference images.
%

% \begin{figure*}[!t]
%     \centering
%     \includegraphics[width= \textwidth,valign=t]{images/mixed_expert_model_1102.png}\vspace{-0.6em}
%     \caption{Visual results of mixed expert model ablation study on real-world video colorization}
%     \label{fig:mixed_expert_model}
%     \vspace{-1em}
% \end{figure*}

\noindent \textbf{Effectiveness of the mixed expert block.}
The MEB is used to explore image prior such as semantic information and object edges to guide the colorization of areas at the boundaries between adjacent objects.
To demonstrate the effectiveness of MEB, we compare with the baselines by replacing the segmentation mask $m_t^{s}$ and the edge mask $m_t^{e}$ one by one.
We also conduct an experiment to remove both $m_t^{s}$ and $m_t^{e}$.
Table~\ref{tab:ablation} shows that the colorization model without semantic image prior does not colorize the videos well.
Figure~\ref{fig:mixed_expert_model} shows the effectiveness of the proposed MEB visually.
We note that the \xnet without using the $m_t^{s}$ or the $m_t^{e}$ does not colorize the frame well where the results contain significant color-bleeding artifacts. For example, the colors of the T-shirt spread to the arm.
In constraint, using the proposed MEB is able to reduce this phenomenon and generates a better-colorized frame (Figure~\ref{fig:mixed_expert_model}(g)).
% utilize temporal clues to select well-colorized regions. It weights pixels transferred from reference images to achieve frame-specific feature fusion. To demonstrate its effectiveness, we replace the TFB with average weighting to fuse the bidirectional features. According to Table~\ref{tab:ablation}, our model without TFB does not perform as well as our complete model.
% Next, we remove the mixed expert block one-by-one to show its significance. Without such crucial clues, the network could not distinguish the edges or semantic margins, as shown in Figure~\ref{fig:mixed_expert_model}, thereby failed to restore the structured degradations. Compared with the baseline, which uses both the $m_t^{s}$ and the $m_t^{e}$ to extract image proirs, the metrics like PSNR and SSIM drops a lot. Besides, we also show its qualitative result in Figure~\ref{fig:mixed_expert_model}, where the arms of the dancer are not handled well without the help of the those masks.

% 1106 multiviewer bike_1106
% camel
\begin{figure}[!tb]
	\setlength\tabcolsep{1.0pt}
	\centering
	\small
	\begin{tabularx}{0.5\textwidth}{cccc}
        \includegraphics[width=0.112\textwidth, height = 0.1\textwidth]{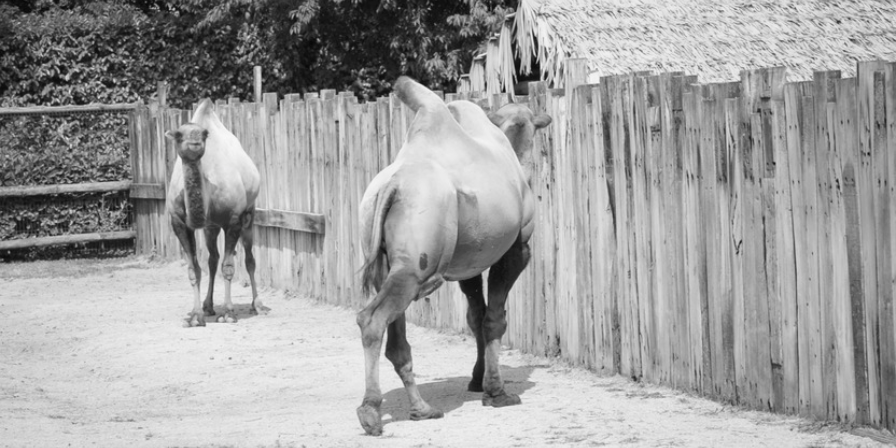}
        \includegraphics[width=0.112\textwidth, height = 0.1\textwidth]{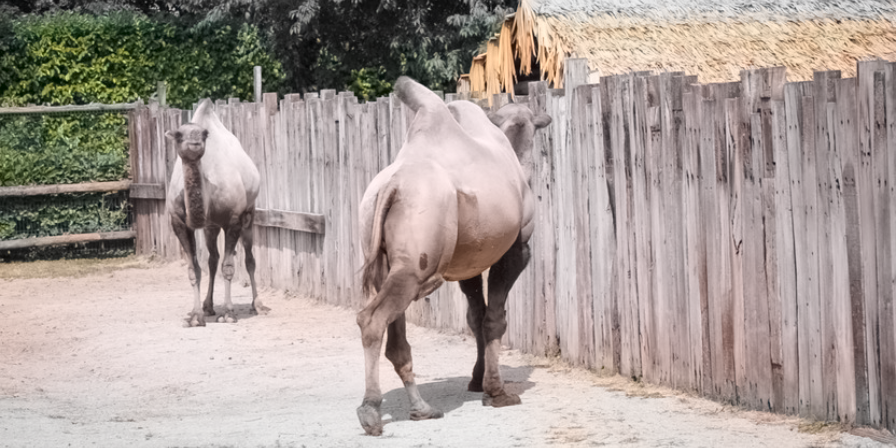}
        \includegraphics[width=0.112\textwidth, height = 0.1\textwidth]{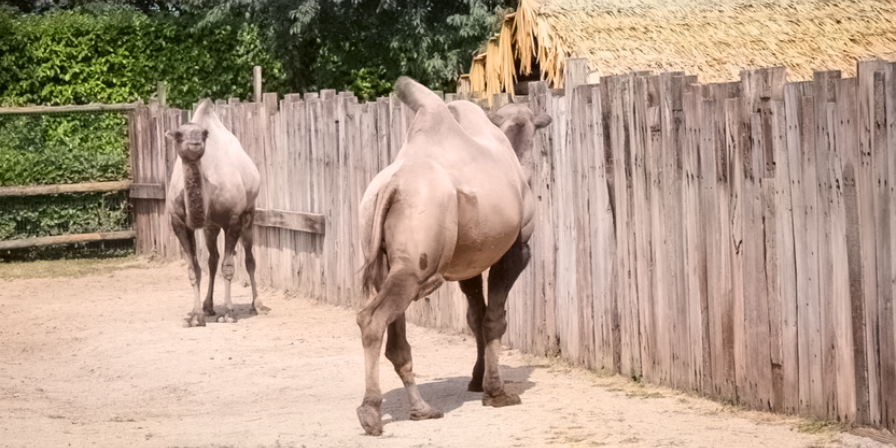}
        \includegraphics[width=0.112\textwidth, height = 0.1\textwidth]{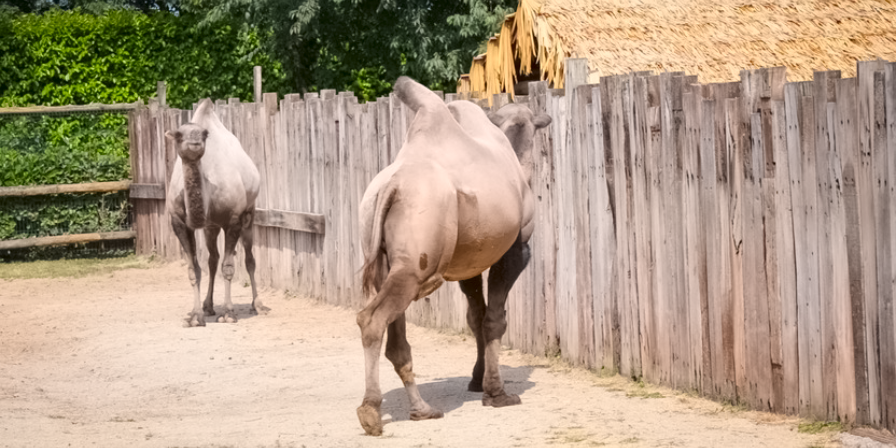}
        \\
        \makebox[0.110\textwidth]{\small (a) Input}
        \makebox[0.110\textwidth]{\small (b) w/o MSRB}
        \makebox[0.110\textwidth]{\small (c) Ours}
        \makebox[0.110\textwidth]{\small (d) Ground Truth} \\ %& \vspace{-0.7em}

	\end{tabularx}
	\vspace{-0.7em}
	\caption{{Effectiveness of the proposed MEB module. As shown in (b), the result by the method without using the MSRB contains uneven color distribution from left to right. The method using the MSRB restores the better-colorized frame in (c)}.}
	\label{fig:abl_srb}
	\vspace{-5mm}
\end{figure}

\noindent \textbf{Effectiveness of the multi-scale recurrent block.}
The MSRB is based on the coarse-to-fine strategy that is widely used in image and video restoration~\cite{Pan_2020_CVPR, gopro2017}. However, its effect is not clear in video coloration.
To validate its effectiveness on video colorization, we compare the proposed method without this block.
The comparisons of ``Ours w/o MSRB" and ``Ours" in Table~\ref{tab:ablation} show that using the MSRB is able to improve the performance of video colorization.
Specifically, as shown in Figure~\ref{fig:abl_srb}, the global coloring effect is not good and the color of the fence is contaminated by the color of the roof around the right area of the image.
The model with the proposed MSRB generates a better-colorized frame (Figure~\ref{fig:abl_srb}(c)).

\noindent \textbf{Effectiveness of the edge loss and hard example mining loss.}
We further examine the effect of the edge loss and the hard example mining on video colorization.
The comparisons of the results in Table~\ref{tab:ablation} (see ``Ours w/o $\mathcal{L}_{edge}$", ``Ours w/o $\mathcal{L}_{hem}$", ``Ours w/o $\mathcal{L}_{edge}$\&$\mathcal{L}_{hem}$", and ``Ours") show that using these two loss functions helps video colorization.

%Recently, the edge loss $\mathcal{L}_{edge}$ is used in interactive image colorization task~\cite{Kim_2021_ICCV} to enhance the edges in a target region. However, its effect is not clear in video coloration. To validate the effectiveness of $\mathcal{L}_{edge}$, we build a baseline model with $\lambda_{edge}=0$. $\lambda_{edge}$ is the parameter in the $\mathcal{L}_{edge}$. The sixth row in Table~\ref{tab:ablation} shows that using $\mathcal{L}_{edge}$ is able to improve the performance of video colorization.

%The hard example mining loss $\mathcal{L}_{hem}$ in a common strategy to dig for hard samples in image and video restoration~\cite{Pan_2020_CVPR, Xu_2019_CVPR}. We conduct an ablation study to validate the effectiveness in video colorization. We build a baseline model with $\lambda_{hem}=0$. $\lambda_{hem}$ is the parameter in $\mathcal{L}_{hem}$. As shown in the seventh row in Table~\ref{tab:ablation}, colorization model without $\mathcal{L}_{hem}$ is much inferior to ours \xnet. We also conduct an experiment to remove both $\mathcal{L}_{edge}$ and $\mathcal{L}_{hem}$. The result is shown in the eighth row in Table~\ref{tab:ablation}.

\noindent \textbf{Two or single reference exemplars.}
As  our method uses two reference frames, one may wonder whether the proposed \xnet works well if only a single reference frame is used.
To answer this question, we build a baseline model that only uses the first color frame of a video (\textit{i.e.}, the frame at time $0$) and train this baseline model using the same settings as the proposed method for fair comparisons.
Table~\ref{tab:psnr} shows that although using a single reference does not generate better results compared to the method using double reference frames, it still achieves better performance than the state-of-the-art methods.

\noindent \textbf{Limitations and future work.}
Although the proposed method achieves favorable performance on video colorization, there are still some limitations. As the colors of the old videos are unknown and subjective, only exploring the colors of several exemplars are not enough to colorize videos. How to explore the domain knowledge of the videos and establish their relations with the well-known knowledge for the colors of scenarios would provide more convincing color information for video colorization.

\section{Conclusion}
In this paper, we propose an effective \xnet to better explore and propagate colors from reference exemplars for video colorization.
We first establish the semantic correspondence between each frame and the reference exemplars in a deep feature space and develop a simple yet effective bidirectional temporal fusion block to better propagate the colors of reference exemplars and avoid the inaccurately matched colors from exemplars.
We develop a mixed expert block to guide the colorization of the regions around object boundaries.
By progressively colorizing frames in a coarse-to-fine manner, we show that the proposed \xnet performs favorably against state-of-the-art methods on the benchmark datasets.

% \vspace{0.2em}
% \noindent \textbf{Acknowledgements.} Ming-Hsuan Yang is supported by the NSF CAREER grant 1149783. Munawar Hayat is supported by the ARC DECRA Fellowship DE200101100. Special thanks to Abdullah Abuolaim and Zhendong Wang for providing the results.

%%%%%%%%% REFERENCES
{\small
\bibliographystyle{ieee_fullname}
\bibliography{bib}
}

\end{document}